%% file: example_paper.tex
\documentclass[nohyperref]{article}

\usepackage{microtype}
\usepackage{graphicx}
\usepackage{booktabs} % for professional tables

\usepackage{hyperref}

\usepackage[accepted]{icml2022}

\usepackage{amsmath}
\usepackage{amssymb}
\usepackage{mathtools}
\usepackage{amsthm}

\newcommand{\E}[3][]{\mathbb{E}_{#2}#1[#3#1]}
\newlength\widthE
\newcommand{\Ebelow}[3][]{\settowidth\widthE{$\operatorname{E}$} \mathop{\mathbb{E}}_{\vphantom{|^|}\mathmakebox[0.5\widthE][l]{#2}}#1[#3#1]}
\usepackage{caption}
\usepackage{subcaption}
\usepackage[capitalize,noabbrev]{cleveref}

\theoremstyle{plain}
\newtheorem{theorem}{Theorem}[section]

\theoremstyle{definition}

\theoremstyle{remark}

\usepackage[textsize=tiny]{todonotes}

\icmltitlerunning{Flow-based Recurrent Belief State Learning for POMDPs}

\begin{document}

\twocolumn[
\icmltitle{Flow-based Recurrent Belief State Learning for POMDPs}




\begin{icmlauthorlist}
\icmlauthor{Xiaoyu Chen}{yyy}
\icmlauthor{Yao Mu}{comp}
\icmlauthor{Ping Luo}{comp}
\icmlauthor{Shengbo Li}{zzz}
\icmlauthor{Jianyu Chen}{yyy,sch}
\end{icmlauthorlist}

\icmlaffiliation{yyy}{Institute for Interdisciplinary Information Sciences, Tsinghua University}
\icmlaffiliation{sch}{Shanghai Qizhi Institute}
\icmlaffiliation{zzz}{School of Vehicle and Mobility, Tsinghua University}
\icmlaffiliation{comp}{Department of Computer Science, The University of Hong Kong}

\icmlcorrespondingauthor{Jianyu Chen}{jianyuchen@tsinghua.edu.cn}

\icmlkeywords{Machine Learning, ICML}

\vskip 0.3in
]



\printAffiliationsAndNotice{}  % leave blank if no need to mention equal contribution

\begin{abstract}
Partially Observable Markov Decision Process (POMDP) provides a principled and generic framework to model real world sequential decision making processes but yet remains unsolved, especially for high dimensional continuous space and unknown models. 
The main challenge lies in how to accurately obtain the belief state, which is the probability distribution over the unobservable environment states given historical information. Accurately calculating this belief state is a precondition for obtaining an optimal policy of POMDPs.
Recent advances in deep learning techniques show great potential to learn good belief states.
However, existing methods can only learn approximated distribution with limited flexibility.
% but they assume the belief states follow certain types of simple distributions such as diagonal Gaussian, which imposes strong restrictions to precisely capture the real belief states.
In this paper, we introduce the \textbf{F}l\textbf{O}w-based \textbf{R}ecurrent \textbf{BE}lief \textbf{S}tate model (FORBES), which incorporates normalizing flows into the variational inference to learn general continuous belief states for POMDPs. Furthermore, we show that the learned belief states can be plugged into downstream RL algorithms to improve performance. In experiments, we show that our methods successfully capture the complex belief states that enable multi-modal predictions as well as high quality reconstructions, and results on challenging visual-motor control tasks show that our method achieves superior performance and sample efficiency.
\end{abstract}

\input{src/1-intro}
\input{src/2-settings}
\input{src/3-method}
\input{src/4-exp}
\input{src/5-survey}

\input{src/6-conclusion}
\nocite{langley00}

\bibliography{example_paper}
\bibliographystyle{icml2022}

%%%%%%%%%%%%%%%%%%%%%%%%%%%%%%%%%%%%%%%%%%%%%%%%%%%%%%%%%%%%%%%%%%%%%%%%%%%%%%%
%%%%%%%%%%%%%%%%%%%%%%%%%%%%%%%%%%%%%%%%%%%%%%%%%%%%%%%%%%%%%%%%%%%%%%%%%%%%%%%
% APPENDIX
%%%%%%%%%%%%%%%%%%%%%%%%%%%%%%%%%%%%%%%%%%%%%%%%%%%%%%%%%%%%%%%%%%%%%%%%%%%%%%%
%%%%%%%%%%%%%%%%%%%%%%%%%%%%%%%%%%%%%%%%%%%%%%%%%%%%%%%%%%%%%%%%%%%%%%%%%%%%%%%
\newpage
\appendix
\onecolumn
\input{src/7-sup}

% You can have as much text here as you want. The main body must be at most $8$ pages long.
% For the final version, one more page can be added.
% If you want, you can use an appendix like this one, even using the one-column format.
%%%%%%%%%%%%%%%%%%%%%%%%%%%%%%%%%%%%%%%%%%%%%%%%%%%%%%%%%%%%%%%%%%%%%%%%%%%%%%%
%%%%%%%%%%%%%%%%%%%%%%%%%%%%%%%%%%%%%%%%%%%%%%%%%%%%%%%%%%%%%%%%%%%%%%%%%%%%%%%

\end{document}

%% file: src/1-intro.tex
\section{Introduction}

% part1. introduce to the POMDP, belief state and the problem.

Partially Observable Markov Decision Process (POMDP) \citep{ASTROM1965174} provides a principled and generic framework to model real world sequential decision making processes.
%but yet remains unsolved, especially for high dimensional continuous {\color{blue} state} space.
Unlike Markov Decision Process (MDP), the observations of a POMDP are generally non-Markovian. Therefore, to make optimal decisions, the agent needs to consider all historical information, which is usually intractable. 
One effective solution is to obtain the belief state.
%, which
The belief state is defined as the probability distribution of the unobservable environment state conditioned on the past observations and actions \citep{kaelbling1998planning}. Such belief state accurately summarizes the history.
% Traditional methods
Traditional methods of calculating belief states \citep{smallwood1973optimal, sondik1971optimal, kaelbling1998planning} assume finite discrete space with a known model. % and derive the belief update with an explicit update. These methods work well in the relatively simple model but suffer from the curse of dimensionality \citep{kaelbling1998planning}. 
In many real world problems, however, the underlying model remains unknown, and the state space is large and even continuous.
To track belief states in POMDPS with continuous state and action spaces, another line of works \citep{thrun1999monte, silver2010monte} uses Monte Carlo algorithms like particle filters to estimate belief states. 
%by sets of samples drawn from belief states.
%More recently,
\citet{nishiyama2012hilbert} proposes to solve the POMDP based on models defined in appropriate reproducing kernel Hilbert spaces (RKHSs). However, this requires access to samples from hidden states during training. 
% However, given the increasingly complex model and high-dimensional state, how to accurately obtain the belief state becomes a hindrance on solving POMDPs.
% Accurately calculating belief states become a necessity for the agent to further obtain an optimal policy.

% part2. problems in existing methods.
% World Model
With the recent advances of deep learning technologies, recent works mainly focus on POMDPs with unknown models and continuous state spaces.
% a branch of works have been proposed to learn the belief states of POMDPs with unknown model and continuous state space.  % with the help of Recurrent Neural Network 
%\citep{krishnan2015deep, gregor2019temporal, lee2020stochastic, hafner2019learning, hafner2019dream, hafner2021mastering}.
%ha2018world does not learn the so defined belief state of POMDP, let's put it in related  works.
To capture belief states, a branch of works including \citet{hausknecht2015deep, gregor2019shaping} uses vector-based representations, namely scalars, to represent belief states. 
However, vector-based belief states may fall short in making predictions for multiple future trajectories (as discussed in Appendix \ref{append-relatedworks}).
Another line of works proposes to learn belief states by approximating belief state distributions.
% For instance, \cite{ma2020particle,igl2018deep} uses a deep version of particle filter to estimate the belief states with samples. {\color{green} Should we attack it here or leave to Section 5?} {\color{blue} (However, particle filter based methods suffer from the curse of dimensionality.) }}
The current state-of-the-art performance on many visual-motor control tasks is also achieved in this manner by sequentially maximizing the observation probability at each timestep using the variational inference \citep{hafner2019dream, zhu2020bridging, okada2020planet, ma2020contrastive}.
They approximate the belief states with distributions like diagonal Gaussians \citep{krishnan2015deep, han2019variational, gregor2019temporal, hafner2019learning, hafner2019dream, lee2020stochastic}, Gaussian mixture \cite{tschiatschek2018variational}, categorical distribution \cite{hafner2021mastering}, or particle filters \cite{ma2020particle,igl2018deep}.
%% Let's not restrict our vision to RSSM
%Among these works, Recurrent State Space Model (RSSM) \citep{hafner2019learning} is a promising method to learn the belief state, and has achieved the state-of-the-art performance on many visual control tasks, and has been widely accepted by \citet{hafner2019dream, zhu2020bridging, okada2020planet, ma2020contrastive}, etc.
%RSSM proposed to solve the belief state learning problem by sequentially maximizing the observation probability at each timestep
%given the historical information
% using the variational inference. 
 %\cite{hafner2019learning, hafner2020dream}. 
However, they still cannot capture general belief states due to the intractability of complex distributions in high-dimensional continuous space. They either suffer from the curse of dimensionality, or instead make some assumptions and learn only the approximated distributions.
Those approximation imposes strong restrictions and is problematic.
% According to \citet{hafner2021mastering}, the Gaussian assumptions make it difficult to predict multi-modal changes between one image and the next. 
% Also, the reconstruction quality is also challenged which hints the representations can be further improved \citep{okada2021dreaming}.
%Variational inference with Gaussian assumptions has been reported to suffer from the posterior intersecting and collapseissue \citet{tolstikhin2019wasserstein}. We believe that a similar problem exists in belief learning. 

\begin{figure}[t]
\vskip 0.2in
\begin{center}
\centerline{\includegraphics[width=\columnwidth]{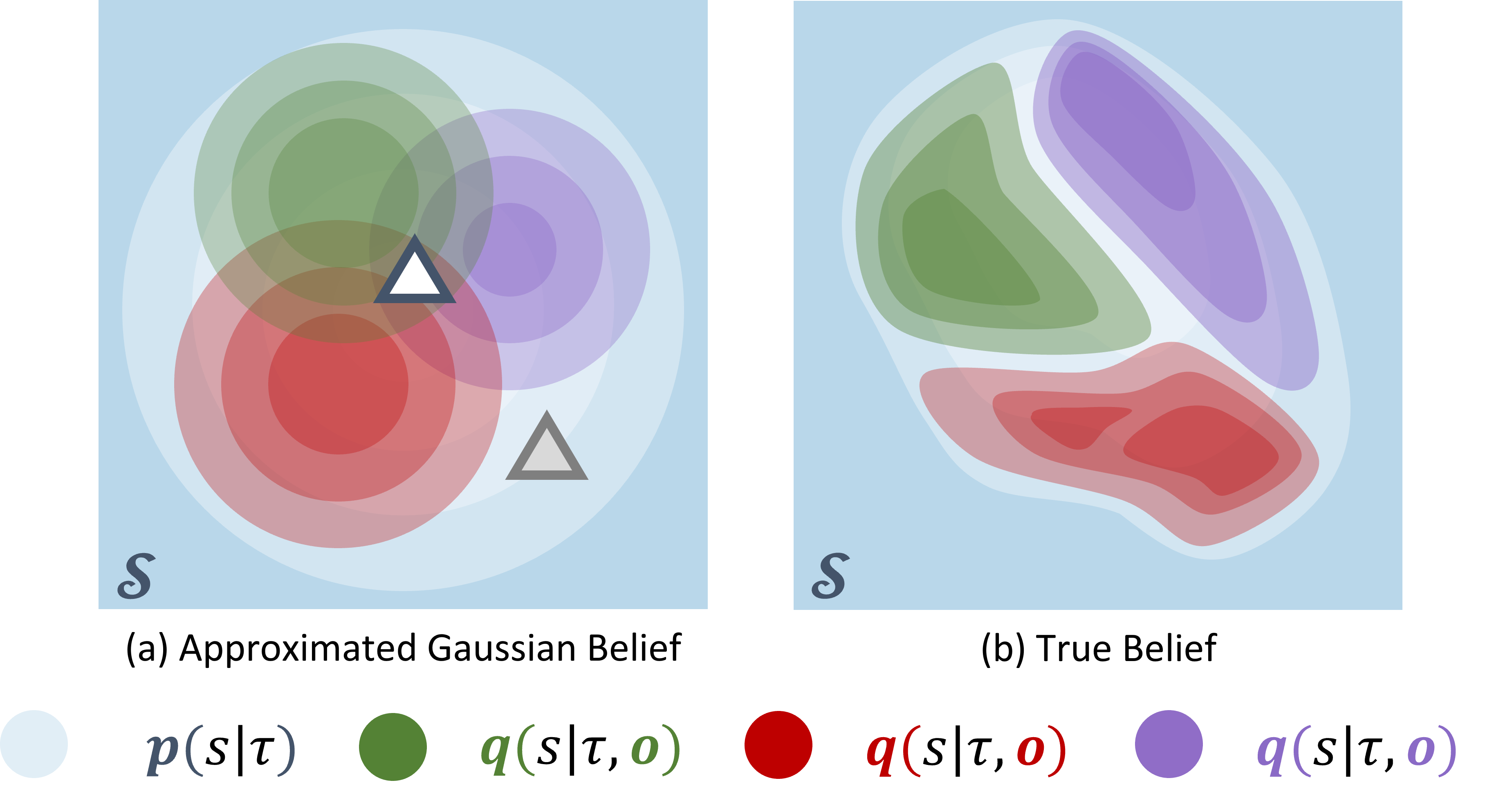}}
\caption{Difference between (a) spherical Gaussian belief states and (b) true belief states (better viewed in color). The spherical Gaussian belief states (a) approximate the true belief states in (b) using Gaussian assuptions, which may result in intersection (the white triangle) or vacancy (the gray triangle) points in the state space.}
\label{latent}
\end{center}
\vskip -0.2in
\end{figure}

Taking the Gaussian belief assumption as an example:
as shown in Figure \ref{latent}, the blue area denotes the unobservable state space of the POMDP. Given the past information $\tau$, the agent maintains a prior distribution of the state $s$, denoted as $p(s|\tau)$ (the distribution in white). Each colored distribution corresponds to the belief state after receiving a different new observation $o$, denoted as the posterior distribution $q(s|\tau,o)$.
Consider an example of the true beliefs as shown in Figure \ref{latent}(b),  %The variational lower bound pulls the beliefs to match the prior belief. % However, the spherical shape of Gaussian may meet troubles when fitting the prior, which ends up in intersecting (the white rectangle part).
with their Gaussian approximations shown in Figure \ref{latent}(a).
The approximation error of Gaussian distributions will easily result in problems of intersecting belief which leads to a mixed-up state (e.g., the white triangle), and empty belief, which leads to a meaningless state (e.g., the grey triangle).
% The corresponding states will be collapsed and mixed up.
This also explains the poor reconstruction problems in interactive environments observed by \citet{okada2021dreaming}.
Furthermore, as mentioned in \citet{hafner2021mastering}, the Gaussian approximation of belief states also makes it difficult to predict multi-modal future behaviours.
Therefore, it is preferable to relax the Gaussian assumptions and use a more flexible family of distributions to learn accurate belief states as shown in Figure \ref{latent}(b). For a more detailed discussion of the related works, please check Section \ref{sec:relatedwork} and Appendix \ref{append-relatedworks}.

% \begin{figure}[t]
%     \centering
%     \includegraphics[width=\linewidth]{figs/latent.png}
%     \caption{ Difference between (a) spherical Gaussian belief states and (b) true belief states (Better viewed in color). }
%     \label{latent}
% \end{figure}

% , the flexibility of distributions allows the belief to be disjoint.
% is the one that is highly flexible, and preferably flexible enough to contain the true belief distribution. 

% \begin{figure}[ht]
%     \begin{subfigure}{.3\textwidth}
%         \centering
%         \includegraphics[width=.7\linewidth]{iclr2022/figs/latent1.jpg}
%         \subcaption{Gaussian Belief}
%         \label{latent-1}
%     \end{subfigure}
%     \begin{subfigure}{.45\textwidth}
%         \centering
%         \includegraphics[width=.7\linewidth]{iclr2022/figs/latent2.jpg}
%         \subcaption{True Belief}
%         \label{latent-2}
%     \end{subfigure}

%     \centering
%     \caption{ Difference between spherical Gaussian belief states \ref{latent-1} and true belief states \ref{latent-2} (Better viewed in color). }
%     \label{latent}
% \end{figure}

% part3. insight, our methods details, tricks, results

In this paper, we propose a new method called \textbf{F}l\textbf{O}w-based \textbf{R}ecurrent \textbf{BE}lief \textbf{S}tate model (FORBES) that is able to learn general continuous belief states for POMDPs. FORBES incorporates Normalizing Flows \citep{tabak2013family, rezende2015variational, dinh2017density} into the variational inference step to construct flexible belief states. In experiments, we show that FORBES allows the agent to maintain flexible belief states, which result in multi-modal and precise predictions as well as higher quality reconstructions. We also demonstrate the results combining FORBES with downstream RL algorithms on challenging visual-motor control tasks (DeepMind Control Suite, \cite{tassa2018deepmind}). The results show the efficacy of FORBES in terms of improving both performance and sample efficiency.

Our contributions can be summarized as follows: 
\begin{itemize}
    \item We propose FORBES, the first flow-based belief state learning algorithm that is capable of learning general continuous belief states for POMDPs.
    
    \item We incorporate FORBES into a POMDP RL framework for visual-motor control tasks that can fully exploit the benefits brought by FORBES.
    
    \item Empirically, %by simpling switching to FORBES and applying small modifications {\color{blue} how to mention the tricks here}, 
    we show that FORBES allows the agent to learn flexible belief states that enable multi-modal predictions as well as high quality reconstructions and help improve both performance and sample efficiency for challenging visual-motor control tasks.
\end{itemize}

%% file: src/2-settings.tex
% \vspace{-7pt}
\section{Preliminaries}

% \vspace{-4pt}
\subsection{Partially Observable Markov Decision Process} 
% $(O,S,A,P,R,\gamma)$.
%A discrete-time POMDP models the relationship between an agent and its environment. 
% \vspace{-4pt}

\begin{figure}[h]
    \centering
    \includegraphics[width=0.8\linewidth]{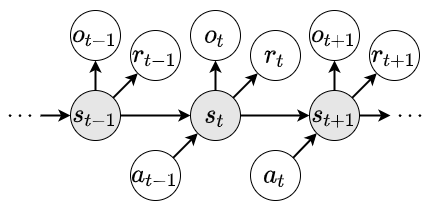}
    \caption{ The PGM of POMDP. The grey circle represents the unobservable hidden states $s$, while the observations $o$, rewards $r$ are observable, and the actions $a$ are determined by the agent.}
    \label{PGM-POMDP}
\end{figure}

Formally, a Partially Observable Markov Decision Process (POMDP) is a 7-tuple $(\mathcal{S}, \mathcal{A}, T, R, \Omega, O, \gamma)$, where $ \mathcal{S} $ is a set of states, $ \mathcal{A} $ is a set of actions, $ T $ is a set of conditional transition probabilities between states, $ R $ is the reward function, $\Omega $ is a set of observations, $ O $ is a set of conditional observation probabilities, and $\gamma $ is the discount factor.

At each timestep $t-1$, the state of the environment is $s_{t-1} \in \mathcal{S}$. The agent takes an action $a_{t-1} \in \mathcal{A}$, which causes the environment to transit to state $s_t$ with probability $T\left(s_t \mid s_{t-1}, a_{t-1}\right)$. The agent then receives an observation $o_t\in \Omega$ which depends on the new state of the environment $s_t$ with probability $O\left(o_t \mid s_t\right)$. Finally, the agent receives a reward $r_{t-1}$ equal to $R(s_{t-1})$. 
The agent's goal is to maximize the the expected sum of discounted rewards $\mathbb{E}\left[\sum_{t=0}^{\infty} \gamma^t r_t \right]$.
Such a POMDP model can also be described as a probabilistic graphical model (PGM) as shown in Figure \ref{PGM-POMDP}.
After having taken action $a_{t-1}$ and observing $o_t$, an agent needs to update its belief state, which is defined as the probability distribution of the environment state conditioned on all historical information:% denoted as $b(s_t)$. According to the conditional independence defined in the PGM shown in Figure \ref{PGM-POMDP}, the belief can be calculated as:
\begin{equation}
\begin{aligned}
    b(s_t)&=p(s_t\mid \tau_t, o_t)\\
    % &=\int_{s_t} p(s_{t+1}\mid s_t, a_t, o_{t+1})b(s_t)
\end{aligned}
\label{eq:belief-def}
\end{equation}
where $\tau_t = \{o_1, a_1, \ldots, o_{t-1}, a_{t-1}\}$.

\subsection{Normalizing Flow}
% part1. introduction to normalizing flow
Instead of using the Gaussian family to approximate the prior and posterior belief distributions, we believe it is more desirable to use a family of distributions that is highly flexible and preferably flexible enough to describe all possible true belief states. Therefore, we use Normalizing Flows \citep{tabak2013family,rezende2015variational} to parameterize those distributions.

Rather than directly parameterizing statistics of the distribution itself, Normalizing Flows model the transformations, or the “flow” progress, needed to derive such a distribution. 
More specifically, it describes a sequence of invertible mappings that gradually transform a relatively simple probability density to a more flexible and complex one.

% part2. details 
Let $f_\theta: \mathbb{R}^{D} \rightarrow \mathbb{R}^{D}$ to be an invertible and differentiable mapping in state space parameterized by $\theta$. Given a random variable $\mathbf{x} \in \mathbb{R}^{D}$ with probability distribution $p(\mathbf{x})$, we can derive the probability of the transformed random variable $\mathbf{z} = f_{\theta}(\mathbf{x})$ by applying the change of variable formula:
\begin{align}
    p(\mathbf{z}) &= p(\mathbf{x}) \left| \det \frac{\partial f_{\theta}^{-1}}{\partial \mathbf{z}}  \right| \\
    \log p(\mathbf{z}) &= \log p(\mathbf{x}) - \log \left| \det \frac{\partial f_{\theta}}{\partial \mathbf{z}} \right|
\end{align}
% To make the calculation of the Jacobian determinant practible and able to scale to high dimensions. Therefore, the flow should be required for low-cost computation of the determinant, e.g., the Jacobian should be triangular. 
To construct a highly flexible family of distributions, we can propagate the random variable at beginning $\mathbf{z}_0$ through a sequence of $K$ mappings and get $\mathbf{z}_K = f_{\theta_K} \circ  f_{\theta_{K-1}} \circ \cdots \circ f_{\theta_1} (\mathbf{z}_0)$ with the probability 
\begin{equation}
    \log p_K(\mathbf{z}_K) = \log p(\mathbf{z}_0) - \sum_{k=1}^{K} \log \left| \det \frac{\partial f_{\theta_{k}}}{\partial \mathbf{z}_{k-1} } \right|
\end{equation}
Given a relatively simple distribution of $\mathbf{z}_0$, say, Gaussian distribution, by iteratively applying the transformations, the flow is capable of representing a highly complex distribution with the probability that remains tractable. The parameters $\theta_1,\ldots,\theta_K$ determine the transformations of the flow. 
% To make the flow capable of making different transformations at different historical context, the parameters $\theta_t$ at timestep $t$ should be conditioned on the past trajectory $\tau_t$.

An effective transformation that is widely accepted is affine coupling layer \citep{dinh2017density, kingma2018glow, kingma2017improving}. Given the input $\mathbf{x} \in \mathbb{R}^{D}$, let $s$ and $t$ stand for scale and translation functions which are usually parameterized by neural networks, where $s,t: \mathbb{R}^{k} \rightarrow \mathbb{R}^{D-k}, k<D$. The output, $\mathbf{y}$, can be viewed as a concatenation of its first $k$ dimensions $\mathbf{y}_{1:k}$ and the remaining part $\mathbf{y}_{k+1:D}$:
\begin{align}
    \mathbf{y}_{1:k} &= \mathbf{x}_{1:k}, \nonumber \\
    \mathbf{y}_{k+1:D} &= \mathbf{x}_{k+1:D} \odot \mathrm{exp}(s(\mathbf{x}_{1:k})) + t(\mathbf{x}_{1:k})
\end{align}
where $\odot$ denotes the element-wise product (see details about affine coupling layer in Appendix \ref{affine details}).

%% file: src/3-method.tex
\section{Flow-based Recurrent Belief State Learning}

\subsection{Flow-based Recurrent Belief State model}
\label{sec:method-1}

We propose the \textbf{F}l\textbf{O}w-based \textbf{R}ecurrent \textbf{BE}lief \textbf{S}tate model (FORBES) which learns general continuous belief states via normalizing flows under the variational inference framework.
% incorporates normalizing flows into the variational inference of the deep recurrent state-space models.
% The FORBES model aims to learn accurate belief states from the history of visual observations via normalizing flows and models the latent dynamics by LSTM-like recurrent networks.  
Specifically, the FORBES model consists of components needed to construct the PGM of POMDP as shown in Figure \ref{PGM-POMDP}:
\begin{equation}
\label{Eq:models}
\begin{aligned}
\mathrm{State\ transition\ model:}\quad       &p(s_t|s_{t-1},a_{t-1})\\
\mathrm{Observation\ model:}\quad       &p(o_t|s_t)\\
\mathrm{Reward\ model:}\quad       &p(r_t|s_t)\\
\end{aligned}
\end{equation}

\vspace{-8pt}
In addition, we have a belief inference model $q(s_t|\tau_t,o_t)$ to approximate the true posterior distribution $p(s_t|\tau_t,o_t)$ as defined in Equation \ref{eq:belief-def}, where $\tau_t = \{o_1, a_1, \ldots, o_{t-1}, a_{t-1}\}$ is the past information. The above components of FORBES can be optimized jointly by maximizing the Evidence Lower BOund (ELBO) \citep{jordan1999viintro} or more generally the variational information bottleneck \citep{tishby2000ib,alemi2016vib}:

\vspace{-15pt}

{\small\begin{multline}
\log p(o_{1:T}|a_{1:T}) \\
\begin{aligned}
&\geq\sum_{t=1}^T \Big(
  \E{q(s_t|o_{\leq t},a_{<t})}{\log p(o_t|s_t) + \log p(r_t | s_t)} \quad \\%\eqbr 
  &\quad - \Ebelow[\big]{q(s_{t-1}|\tau_{t-1}, o_{t-1})}{D_{\mathrm{KL}}(q(s_t|\tau_t, o_t) \| p(s_t|s_{t-1},a_{t-1})}) \Big) \doteq \mathcal{J}_{\mathrm{Model}}
\end{aligned}
\label{Eq:Jmodel}
\end{multline}}

\vspace{-5pt}

Detailed derivations can be found in Appendix \ref{append-ELBO}. In practice, the state transition model, observation model, reward model, and belief inference model can be represented by stochastic deep neural networks parameterized by $\psi$:

\vspace{-10pt}
\begin{equation}
\begin{aligned}
    p_\psi(s_t|s_{t-1},a_{t-1}),\  p_\psi(o_t|s_t),\ p_\psi(r_t|s_t), \  q_\psi(s_t|\tau_t,o_t) \nonumber
\end{aligned}
\end{equation}

where their outputs usually follow simple distributions such as diagonal Gaussians. The parameterized belief inference model $q_\psi(s_t|\tau_t,o_t)$ acts as an encoder that encodes the historical information using a combination of convolutional neural networks and recurrent neural networks. 
% {\color{green}Note that $q$ below is obtained through the factorization of $q(s_{1:T}|o_{1:T}, a_{1:T-1}) = \prod_{t} q(s_t | s_{t-1}, a_{t-1}, o_t)$.}

\begin{figure*}[t]
    \begin{subfigure}{.49\textwidth}
        \centering
        \includegraphics[width=0.95\linewidth]{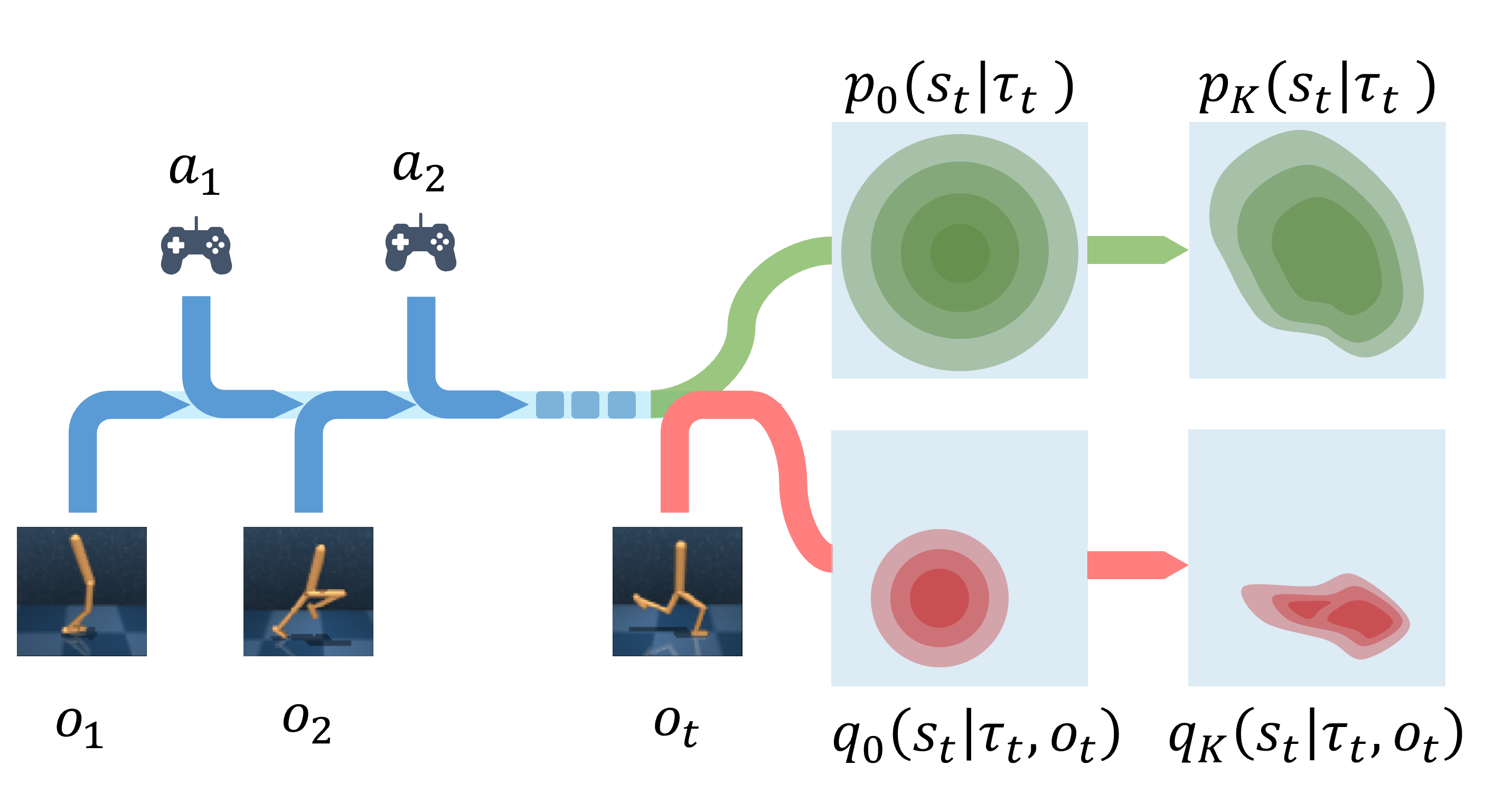}
        \caption{Belief state inference}
        \label{algo-1}
    \end{subfigure}
    \begin{subfigure}{.49\textwidth}
        \centering
        \includegraphics[width=0.95\linewidth]{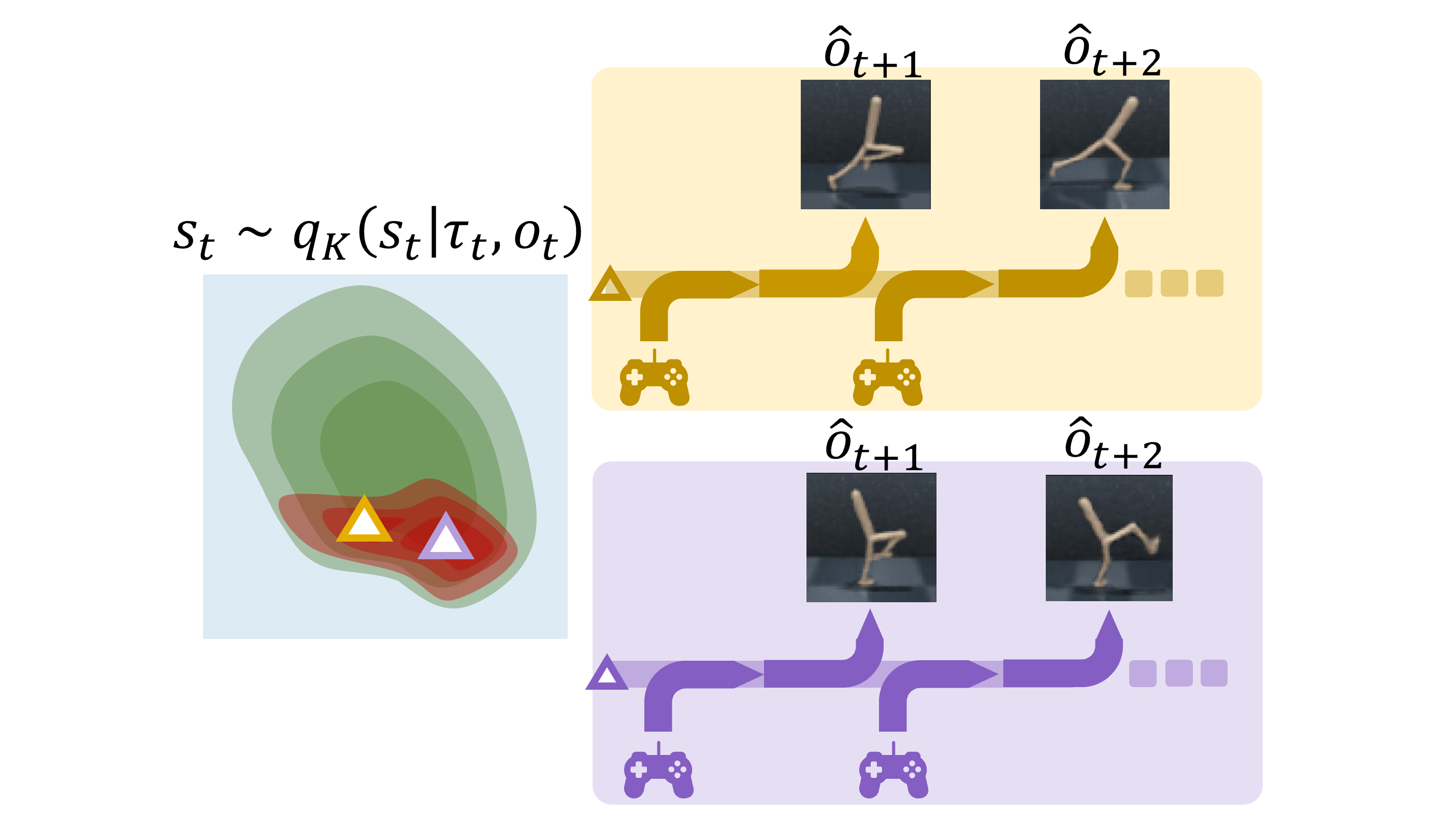}
        \caption{Predictions beginning from different samples}
        \label{algo-2}
    \end{subfigure}
    \centering
    \caption{The algorithm framework of FORBES. Figure \ref{algo-1} shows how to calculate prior and posterior belief distribution given previous information. The blue arrows bring in historical observations and actions, and the green path shows the evolution of prior belief distribution. The red path takes an additional $o_t$ and shows the evolution of posterior belief distribution. Figure \ref{algo-2} shows the predictions of future trajectories starting from different samples (yellow and purple triangles) given the future actions. 
    \label{algo-fig}}
\end{figure*}

In FORBES we provide special treatments for the belief inference model and the state transition model to represent more complex and flexible posterior and prior distributions. As shown in Figure \ref{algo-fig}(a), the input images $o_{1:t}$ and actions $a_{1:t-1}$ are encoded with $q_\psi(s_t|\tau_t,o_t)$ (the blue and the red path). Then our final inferred belief is obtained by propagating $q_\psi(s_t|\tau_t,o_t)$ through a set of normalizing flow mappings denoted $f_{\theta_K} \circ \cdots \circ f_{\theta_1}$ to get a representative posterior distribution $q_{\psi,\theta}(s_t|\tau_t, o_t)$. For convenience, we denote $q_0=q_\psi$ and $q_K=q_{\psi,\theta}$.  On the other hand, $o_{1:t-1}$ and $a_{1:t-2}$ are encoded with $q_\psi(s_{t-1}|\tau_{t-1},o_{t-1})$ (the blue path), then the state transition model is used to obtain the prior guess of the state $p_\psi(s_t\mid \tau_t)=\mathbb{E}_{q_\psi(s_{t-1}|\tau_{t-1},o_{t-1})} \left[p_\psi(s_t\mid s_{t-1},a_{t-1})\right]$ (the green path). Then our final prior is obtained by propagating $p_\psi(s_t|\tau_t)$ through another set of normalizing flow mappings denoted $f_{\omega_K} \circ \cdots \circ f_{\omega_1}$ to get a representative prior distribution $p_{\psi,\omega}(s_t|\tau_t)$. For convenience, we denote $p_0=p_\psi$ and $p_K=p_{\psi,\omega}$. Then as shown in Figure \ref{algo-fig}(b), we can sample the initial state $s_t$ (the yellow and purple triangles) from the belief states $q_K(s_t\mid\tau_t,o_t)$. For each sampled initial state, we can use the state transition model to predict the future states $\hat{s}_{t+h}$ given the future actions $a_{t:t+h-1}$, and then use the observation model to reconstruct the observations $\hat{o}_{t+h}$, where $h$ is the prediction horizon.
With the above settings, we can substitute the density probability inside the KL-divergence term in Equation \ref{Eq:Jmodel} with Normalizing Flow:

\vspace{-22pt}
\begin{equation}
\small
\begin{aligned}
\label{eq:KL}
    \log q_K(s_t|\tau_t, o_t) &= \log q_0(s_t|\tau_t, o_t) - \sum_{k=1}^{K} \log \left| \det \frac{\partial f_{\theta_k}}{\partial s_{t,k-1}} \right| \\
    \log p_K(s_t|\tau_t) &= \log p_0(s_t|\tau_t) - \sum_{k=1}^{K} \log \left| \det \frac{\partial f_{\omega_k}}{\partial s_{t,k-1}} \right| \\
    % &\mathbb{E}_{q_K(s_{t-1}|s_{t-2},a_{t-2},o_{t-1})}\left[ D_{\mathrm{KL}} (q_K(s_t|s_{t-1}, a_{t-1}, o_t) \| p_K(s_t |s_{t-1},a_{t-1}) ) \right] \\
    % &= \mathbb{E}_{q_K(s_{t-1}|s_{t-2},a_{t-2},o_{t-1})} \left[ \mathbb{E}_{q_K(s_t|s_{t-1}, a_{t-1}, o_t)} \left[ \log q_K(s_t|s_{t-1}, a_{t-1}, o_t) - \log p_K(s_t |s_{t-1},a_{t-1}) \right] \right]\\
    % &= \mathbb{E}_{s_{t-1} \sim q_K} \left[ \mathbb{E}_{s_t \sim q_K} \left[ \log q_0(s_t|s_{t-1}, a_{t-1}, o_t) - \sum_{k=1}^{K} \log \left| \det \frac{\partial f_{\theta_k}}{\partial s_{t,k-1}} \right| - \log p_0(s_t |s_{t-1},a_{t-1}) + \sum_{k=1}^{K} \log \left| \det \frac{\partial f_{\omega_k}}{\partial s_{t,k-1}} \right|  \right] \right] 
\end{aligned}
\end{equation}

where $p_K(s_t\mid s_{t-1},a_{t-1})=p_K(s_t\mid \tau_t)$ given the sampled $s_{t-1}$ from $q_K(s_{1:t}|\tau_t,o_t)$. $s_{t,k}$ is the state variable $s_t$ transformed by $k$ layers of normalizing flows, and $s_{t,0}=s_t$.

To further demonstrate the properties of FORBES, we provide the following theorems.

\begin{theorem}
\label{elbo-thm}
The approximation error of the log-likelihood when maximizing the  $\mathcal{J}_{\mathrm{Model}}$ (the derived ELBO) defined in Equation \ref{Eq:Jmodel} is:

\vspace{-10pt}

\begin{equation}
\begin{aligned}\label{error}
    % \log p(o_{1:T}, r_{1:T}|a_{1:T}) - \mathcal{J}_{\mathrm{Model}} = \sum_t  D_{\mathrm{KL}}(q(s_t | s_{t-1}, a_{t-1}, o_t) \| p(s_t | s_{t-1}, a_{t-1}, o_t))
    &\log p(o_{1:T}, r_{1:T}|a_{1:T}) - \mathcal{J}_{\mathrm{Model}} \\
    &= \Ebelow[\big]{q_K(s_{1:T}|o_{1:T},a_{1:T-1})}{
    \Sigma_{t=1}^{T}
    % \sum_{t=1}^T 
    D_{\mathrm{KL}}(q(s_t|\tau_t, o_t) \| p(s_t\mid \tau_t, o_t))}
\end{aligned}
\end{equation}
where $p(s_t\mid \tau_t, o_t)$ denotes the true belief states. %$b(s_{t})$.
\end{theorem}
% \vspace{-20pt}

% % below is the original theorem
% \begin{theorem}
% \label{elbo-thm}
% The approximation error of the log-likelihood when maximizing the  $\mathcal{J}_{\mathrm{Model}}$ (the derived ELBO) defined in Equation \ref{Eq:Jmodel} is:
% \begin{equation}\label{error}
%     % \log p(o_{1:T}, r_{1:T}|a_{1:T}) - \mathcal{J}_{\mathrm{Model}} = \sum_t  D_{\mathrm{KL}}(q(s_t | s_{t-1}, a_{t-1}, o_t) \| p(s_t | s_{t-1}, a_{t-1}, o_t))
%     \log p(o_{1:T}, r_{1:T}|a_{1:T}) - \mathcal{J}_{\mathrm{Model}} = \mathbb{E}_{q_K(s_{1:T}|\tau_T,o_T)} \left[ \sum_{t=1}^T  D_{\mathrm{KL}}(q(s_t | \tau_{t}, o_t) \| p(s_t\mid \tau_t, o_t)) \right]
% \end{equation}
% where $p(s_t\mid \tau_t, o_t)$ denotes the true belief states. %$b(s_{t})$.
% \end{theorem}

% {\color{red} check the consistensy of positive /negative of $\mathcal{J}_model$}

Detailed proofs can be found in Appendix.\ref{sec:proofs}. Theorem \ref{elbo-thm} suggests that, when the learning algorithm maximizes the  $\mathcal{J}_{\mathrm{Model}}$ (the derived ELBO), then the $D_{\mathrm{KL}}$ terms in the right-hand side are minimized, which indicate the KL-divergence between the learned belief states $q(s_t | \tau_{t}, o_t)$ and the true belief states $p(s_t\mid \tau_t, o_t)$. Clearly, if $p(s_t\mid \tau_t, o_t)$ is a complex distribution and $q(s_t | \tau_{t}, o_t)$ is chosen from a restricted distribution class such as diagonal Gaussian, then when the algorithm maximizes the  $\mathcal{J}_{\mathrm{Model}}$ (the derived ELBO), there will still be a potentially large KL-divergence between the learned and the true belief states.
% if $q_K$ can approximate the $p$ with arbitrarily small error, then the $D_{\mathrm{KL}}$ term in the right-hand side can approach zero. 

\begin{figure*}[t!]
    % \vspace{-10pt}
    \centering
    \includegraphics[width=.9\linewidth]{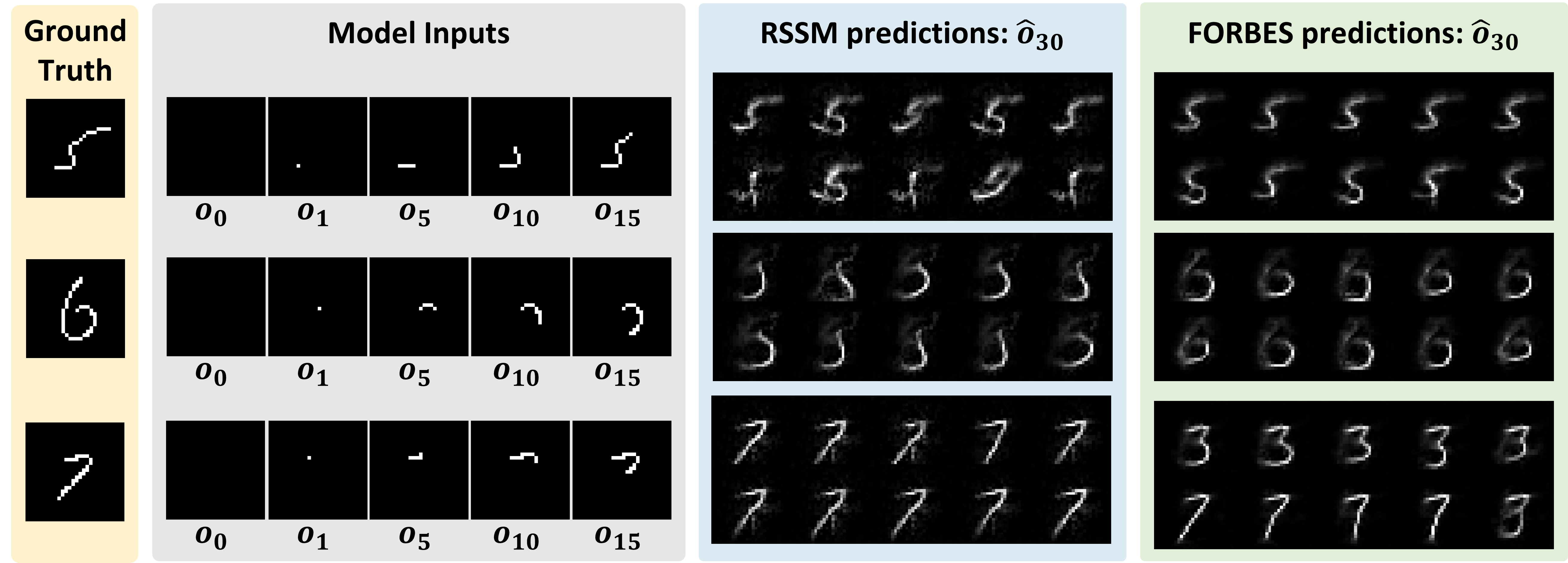}
    
    \centering
    \caption{Predictions on sequential MNIST of two models. This is a digit writing task. The fully written digits are shown in the leftmost column. We use incomplete writing processes (the first 15 frames, partially shown in the grey column) as the inputs and let the models predict the complete digit (as shown in the blue/green columns). The results show that FORBES can make precise yet diverse predictions with less blur and no mode mixup.}
    \label{fig:seq-1}
    \vspace{-10pt}
\end{figure*}

Therefore, naturally there raises the problem that is normalizing flow a universal distributional approximator that is capable of accurately representing arbitrarily complex belief states, so the KL-divergence terms in the right-hand side of Equation (\ref{error}) can be minimized to approach zero? The answer is yes for a wide range of normalizing flows. To be specific, \citet{teshima2020couplingbased} provides theoretical results for the family of the flow used in FORBES.
% transforming a known distribution into any probability measure that we desired? We save the formal definition (Definition \ref{def-dua}) to the appendix. 
% \begin{definition}(Distributional Universal Approximation \citep{teshima2020couplingbased})
% \label{def-dua}
% Let $\mathcal{M}$ be a model which is a set of measurable mappings from $\R^m$ to $\R^n$.
% We say that a model $\mathcal{M}$ is a \emph{distributional universal approximator} if, 
% for any absolutely continuous probability measure $\mu$ on $\R^m$ and any probability measure $\nu$ on $\R^n$, there exists a sequence $\{g_i\}_{i=1}^\infty\subset\mathcal{M}$ such that $(g_i)_*\mu$ converges to $\nu$ in distribution as $i\rightarrow\infty$, where $(g_i)_*\mu:= \mu\circ g_i^{-1}$.
% \end{definition}
% \begin{theorem} (\citet{teshima2020couplingbased})
% Let $\mathcal{M}$ be a set of measurable mappings from $\R^m$ to $\R^m$:
% $\mathcal{M} \doteq \{ W_1 \circ g_1 \circ \cdots W_n \circ g_n: n \in \mathbb{N}, g_i \in \mathcal{G}, W_i \in \mathcal{A}\}$, where $\mathcal{G}$ contains affine coupling layers and $\mathcal{A}$ are affine transforms. Then, for any absolutely continuous probability measure $\mu$ and any probability measure $\nu$ on $\R^m$, there exists a sequence $\{g_i\}_{i=1}^\infty\subset\mathcal{M}$ such that $(g_i)_*\mu$ converges to $\nu$ in distribution as $i\rightarrow\infty$, where $(g_i)_*\mu:= \mu\circ g_i^{-1}$. Namely, $\mathcal{M}$ is a universal distributional approximator.
% \end{theorem}

% In fact, there always exists a diffeomorphism that can turn one well-behaved distribution to another. 
Besides the aforementioned affine coupling flow, many works show the distributional universality of other flows \citep{kong2020expressive, huang2018neural}.
%, which further demonstrates the capacity of other flows.
Ideally, the universal approximation property of the flow model $q_K(s_t\mid \tau_t,o_t)$ allows us to approximate the true posterior $p(s_t\mid \tau_t,o_t)$ with arbitrary accuracy. Thus, compared to previous methods, FORBES helps close the gap between the log-likelihood and the ELBO to obtain a more accurate belief state.
Though we usually cannot achieve the ideal zero KL-divergence in practice, our method can get a smaller approximation error, equally a higher ELBO than previous works. We verify this statement in section \ref{exp:seqMNIST}.

% What's more, as proved in \citet{teshima2020couplingbased, kong2020expressive}, many normalizing flow models, including the one used in FORBES, are universal distributional approximators, which close the gap between the log probability and the ELBO and enable the agent to calculate the probability more acurately.

% {\color{red} should we put this paragraph before thm.1?}
% During the training, the components are optimized jointly to increase the variational lower bound \citep{jordan1999viintro} or more generally the variational information bottleneck  \citep{tishby2000ib,alemi2016vib}, i.e., minimize the $\mathcal{J}_{\mathrm{ELBO}}$.
% \begin{equation}
% \begin{aligned}
% &\mathcal{J}_{\mathrm{ELBO}} = - \mathrm{E}_{p}\left(\sum_{t}\left(\mathcal{J}_{\mathrm{O}}^{t}+\mathcal{J}_{\mathrm{R}}^{t}+\mathcal{J}_{\mathrm{D}}^{t}\right)\right) \quad \mathcal{J}_{\mathrm{O}}^{t} \doteq \ln q\left(o_{t} \mid s_{t}\right) \\
% &\mathcal{J}_{\mathrm{R}}^{t} \doteq \ln q\left(r_{t} \mid s_{t}\right) \quad \mathcal{J}_{\mathrm{D}}^{t} \doteq-\beta D_{\mathrm{KL}}\left(q_K\left(s_{t} \mid s_{t-1}, a_{t-1}, o_{t}\right) \| p_K\left(s_{t} \mid s_{t-1}, a_{t-1}\right)\right)
% \end{aligned}
% \end{equation}

\subsection{ POMDP RL framework based on FORBES }
To show the advantage of the belief states inferred by the FORBES model compared to the existing belief inference method in visual-motor control tasks, we incorporate FORBES into a flow-based belief reinforcement learning algorithm for learning the optimal policy in POMDPs. Inspired by \citet{hafner2019dream}, the algorithm follows an actor-critic framework but is slightly modified to exploit better the flexible nature of FORBES: The critic estimates the accumulated future rewards, and the actor chooses actions to maximize the estimated cumulated rewards. 
Instead of using only one sample, 
both the actor and critic operate on top of the samples of belief states learned by FORBES. They thus benefit from the accurate representations learned by the FORBES model. Note that this is an approximation of the true value on belief, which avoids the intractable integration through observation model. 
% Both the actor and critic operate on top of the learned belief states and thus benefit from the accurate representations learned by the FORBES model. %At each learning iteration, the FORBES model is fixed during the reinforcement learning step, so that the actor and critic gradients do not affect the belief state learning. 

The critic $v_{\xi}\left(s_{\tau}\right)$ aims to predict the discounted sum of future rewards that the actor can achieve given an initial state $s_t$, known as the state value $\mathbb{E}\left(\sum_{\tau=t}^{t+\infty} \gamma^{\tau-t} r_{\tau}\right)$, where $\xi$ denote the parameters of the critic network and $H$ is the prediction horizon. We leverage temporal-difference to learn this value, where the critic is trained towards a value target that is constructed from the intermediate reward and the critic output for the next step's state. In order to trade-off the bias and the variance of the state value estimation, we use the more general TD($\lambda$) target  \citep{sutton2018rlbook}, which is a weighted average of n-step returns for different horizons and is defined as follows:

\vspace{-5pt}
\begin{equation}
V^\lambda_\tau \doteq
\hat{r}_\tau + \hat{\gamma}_\tau
\begin{cases}
  (1 - \lambda) v_\xi(s_{\tau+1}) + \lambda V^\lambda_{\tau+1} & \text{if}\quad \tau<t+H, \\
  v_\xi(s_{t+H}) & \text{if}\quad \tau=t+H. \\
\end{cases}
\end{equation}
To better utilize the flexibility belief states from FORBES, we run the sampling method multiple times to capture the diverse predictions. Specifically, we sample $N$ states from the belief state given by FORBES and then rollout trajectories of future states and rewards using the state transition model and the reward model. Finally, we train the critic to regress the TD($\lambda$) target return using a mean squared error loss:
\begin{equation}\label{Eq:JCritic}
\mathcal{J}_{\mathrm{Critic}}(\xi) = \mathbb{E}\Big[
  \textstyle \sum_{i=1}^{N}
  \textstyle \sum_{\tau=t}^{t+H}
    \frac{1}{2} \big(
      v_\xi(s_{i, \tau}) - \operatorname{sg}(V^\lambda_{i, \tau})
    \big)^2
\Big].
\end{equation}
% \begin{equation}\label{Eq:JCritic}
% \mathcal{J}_{\mathrm{Critic}}(\xi) = 
% \underset{{s_{i,0}\sim q_K, a_\tau\sim q_\phi, s_{i,\tau}\sim p_\psi}}{\mathbb{E}}\Big[
%   \textstyle \sum_{i=1}^{N}
%   \textstyle \sum_{\tau=t}^{t+H}
%     \frac{1}{2} \big(
%       v_\xi(s_{i, \tau}) - \operatorname{sg}(V^\lambda_{i, \tau})
%     \big)^2
% \Big].
% \end{equation}
where $sg(\cdot)$ is the stop gradient operation. The actor $a_{\tau} \sim q_{\phi}\left(a_{\tau} \mid s_{\tau}\right)$ aims to output actions that maximize the prediction of long-term future rewards made by the critic and is trained directly by backpropagating the value gradients through the sequence of sampled states and actions, i.e., maximize:
\begin{equation}\label{Eq:JActor}
\mathcal{J}_{\mathrm{Actor}}(\phi) = 
% \underset{{s_{i,0}\sim q_K, a_\tau\sim q_\phi, s_{i,\tau}\sim p_\psi}}
{\mathbb{E}}\left(\sum_{i=1}^{N} \sum_{\tau=t}^{t+H} \mathrm{~V}^{\lambda}_{i,\tau}\right)
\end{equation}
We jointly optimize the model loss $\mathcal{J}_{Model}$ with respect to the model parameters $\psi$, $\theta$ and $\omega$, the critic loss $\mathcal{J}_{Critic}$ with respect to the critic parameters $\xi$ and the actor $\mathcal{J}_{Actor}$ loss with respect to the actor parameters $\phi$ using the Adam optimizer with different learning rates:
\vspace{-10pt}
\begin{equation}
 \underset{\psi,\xi,\phi,\theta,\omega}{\mathrm{min}} \quad \alpha_{0}\mathcal{J}_{\mathrm{Critc}}(\xi) -\alpha_{1}\mathcal{J}_{\mathrm{Actor}}(\phi) 
 -\alpha_{2}\mathcal{J}_{\mathrm{Model}}(\psi, \theta, \omega)
\label{eq:joint_loss}
\end{equation}
% \begin{equation}
%  \underset{\psi,\xi,\phi,\theta,\omega}{\mathrm{min}} \quad \mathcal{J}_{\mathrm{FORBES}} =  \alpha_{0}\mathcal{J}_{\mathrm{Critc}}(\xi) -\alpha_{1}\mathcal{J}_{\mathrm{Actor}}(\phi) 
%  -\alpha_{2}\mathcal{J}_{\mathrm{Model}}(\psi, \theta, \omega)
% \label{eq:joint_loss}
% \end{equation}
where 
%$\alpha_{0}=5e-4$, $\alpha_{1}=8e-5$, $\alpha_{2}=8e-5$ 
$\alpha_{0}$, $\alpha_{1}$, $\alpha_{2}$  are coefficients for different components, and we summarize the whole framework of optimizing in Algorithm \ref{algo-Forbes}.

%  {\color{red} Notations. A total loss equation like eq5 in bird.}

\begin{algorithm}[tb]
   \caption{FORBES Algorithm}
   \label{algo-Forbes}
\begin{algorithmic}
   \STATE {\bfseries Input:} buffer $\mathcal{B}$, imagination horizon $H$, interacting step $T$, batch size $B$, batch length $L$, number of trajetories $N$.
   \STATE Initialize buffer $\mathcal{B}$ with $S$ random seed episodes.
   \WHILE{ not converged }
            \FOR {$c=1,\dots,C$}
                \STATE Draw $B$ data sequences $\{(o_t,a_t,r_t)\}_{t=k}^{k+L}$ from $\mathcal{B}$ 
                \STATE Infer belief state $q_K(s_t|s_{t-1}, a_{t-1}, o_t)$.
                \FOR {$i=1,\dots,N$}
                    \STATE Rollout imaginary trajectories $\{(s_{i,\tau},a_{i,\tau})\}_{\tau=t}^{t+H}$ with belief transition model.
                \ENDFOR
                \STATE For each $s_{i,\tau}$, predict rewards $p_\psi(r_{i,\tau}|s_{i,\tau})$ and values $v_{\phi}(s_{i,\tau})$  \COMMENT{{\color{gray}\emph{Calculate returns}}}
                \STATE Update $\theta,\omega,\xi,\phi,\psi$ using Equation (\ref{Eq:Jmodel}), (\ref{eq:KL}), (\ref{Eq:JCritic}), (\ref{Eq:JActor}) and (\ref{eq:joint_loss}) \COMMENT{{\color{gray}\emph{Optimize parameters}}}
            \ENDFOR
            \STATE Reset environment and get $o_1$.
            \FOR{$t=1,\dots,T$}
                \STATE Compute $s_t \sim q_K(s_t | s_{t-1}, a_{t-1}, o_t)$ from history.
                \STATE Compute $a_t \sim \pi(a_t|s_t)$ with action model.
                \STATE Add exploration noise to action.
                \STATE Execute $a_t$ and get $o_{t+1}, r_t$.
            \ENDFOR
            \STATE Add experience to buffer $\mathcal{B} = \mathcal{B} \cup \{(o_t,a_t,r_t)_{t=1}^T\}$
        \ENDWHILE
\end{algorithmic}
\end{algorithm}

%% file: src/4-exp.tex
\section{Experiments}

Our experiments evaluate FORBES on two image-based tasks.
%a digit writing task and a challenging visual-motor control task, 
% where FORBES outperforms baselines including RSSM.
We first demonstrate the belief learning capacity on a digit writing task in Section \ref{exp:seqMNIST}, and show that FORBES captures beliefs that allow for multi-modal yet precise long-term predictions as well as higher ELBO.
For large-scale experiments, we test the proposed POMDP RL framework based on FORBES in Section \ref{exp:dmc}. The results of multiple challenging visual-motor control tasks from DeepMind Control Suite \citep{tassa2018deepmind} show that FORBES outperforms baselines in terms of performance and sample efficiency. 
In Section \ref{exp:ablation}, we further provide ablation studies of the multiple imagined trajectories technique used in our method.

\subsection{Digit Writing Tasks} 
\label{exp:seqMNIST}

In this experiment, we validate the capacity of FORBES by modelling the partially observable sequence with visual inputs.
We adopt the MNIST Sequence Dataset \citep{mnist_seq} that consists of sequences of handwriting MNIST digit stokes. This problem can be viewed as a special case of POMDP, whose action space is $\O$ and rewards remain $0$. Such a problem setting separates the belief learning and policy optimizing problem and allows us to concentrate on the former one in this section. We convert the digit stroke to a sequence of images of size $28 \times 28$ to simulate the writing process. At time step $t$, the agent can observe $o_t$ that has already written $t$ pixels, and we train the agent maximizing $\mathcal{J}_{\mathrm{Model}}$ in Equation \ref{Eq:Jmodel} except for the reward reconstruction term. 
% \begin{wrapfigure}{rh}{0.3\textwidth} 
%     \centering
%     \includegraphics[width=0.3\textwidth]{figs/testelbo_.png}
%     \caption{ELBO on digit writing }
%     \label{fig:seq-elbo}
% \end{wrapfigure}
\begin{figure}[h]
    \centering
    \includegraphics[width=0.5\linewidth]{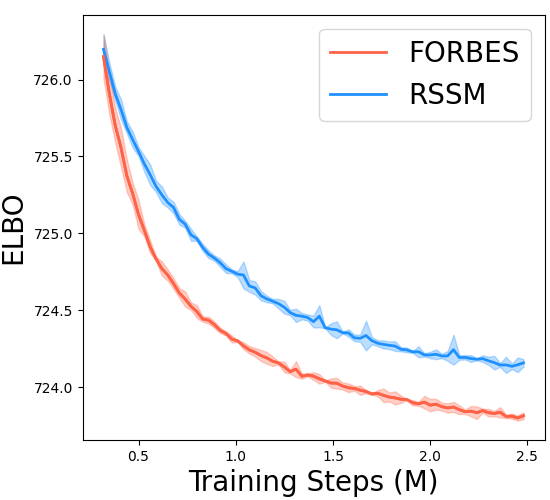}
    \centering
    \caption{ELBO on digit writing.}
    % \vskip -20pt
    % \vspace{-20pt}
    \label{fig:seq-elbo}
\end{figure}
As shown in Figure \ref{fig:seq-1}, we randomly select three digits as examples 
(see Appendix \ref{sup:digit-exp} for more results)
and show the inputs as well as the prediction outputs of our model and the RSSM~\citep{hafner2019learning} baseline, which is the previous state-of-the-art method for learning continuous belief states of POMDPs. 
The leftmost column is the ground truth of the fully written digits. During the testing, we feed the initial 15 frames $\{o_1, o_2, \cdots, o_{15}\}$ to the model, and the columns in grey exhibit a part of the inputs. Then we sample several states from the inferred belief state and rollout via the learned state transition model (Equation (\ref{Eq:models})) for 15 steps and show the reconstruction results of the predictions. 
As shown in the blue and green columns on the right of Figure \ref{fig:seq-1}, though RSSM can also predict the future strokes in general, the reconstructions are relatively blurred and mix different digits up. It also fails to give diverse predictions.
However, FORBES can make precise yet diverse predictions. Each prediction is clear and distinct from other digits. Given the beginning of the digit 7, FORBES successfully predicts both 7 and 3 since they have a similar beginning.
The results can be partially explained via the mixed-up belief and the empty belief as shown in Figure \ref{latent}, which support the claim that FORBES can better capture the complex belief states.

We also provide the quantitative results in Figure \ref{fig:seq-elbo}, which is the $ \mathrm{ELBO}$ on test digits sequence set that is never seen during training. The results show that FORBES can achieve a tighter ELBO, which verifies the theoretical results in \ref{sec:method-1}.
The details of the implementation can be found in Appendix \ref{sec:hparams}.

\vspace{-5pt}
\subsection{Visual-motor control tasks}  %% 2
\label{exp:dmc}

\begin{figure}[t]
    \centering
    % \vspace{-10pt}
    % \includegraphics[width=.7\linewidth]{iclr2022/figs/DMC.pdf}
    \includegraphics[width=\linewidth]{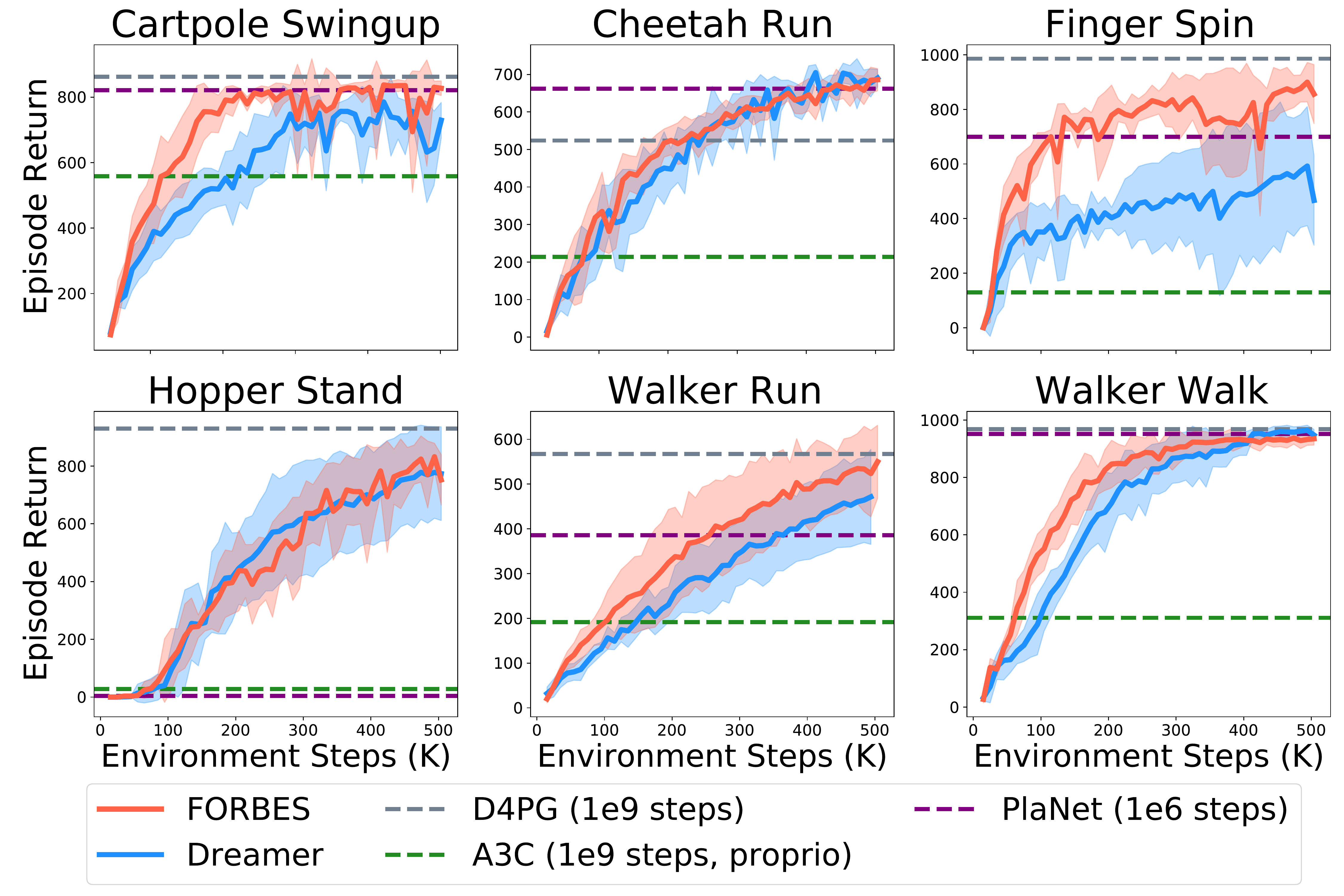}
    \centering
    \caption{Performance on DeepMind Control Suite. The shaded areas show the standard deviation across 5 seeds. FORBES achieves better performance and sample efficiency in various challenging tasks.}
    % \vskip -30pt
    \vspace{-15pt}
    \label{fig:mujoco-exp}

\end{figure}
\vspace{-5pt}
We experimentally evaluate the performance of FORBES on Reinforcement Learning on a variety of visual-motor control tasks from the DeepMind Control Suite \citep{tassa2018deepmind}, illustrated in Figure \ref{fig:mujoco-exp}. Across all the tasks, the observations are $64 \times 64 \times 3$ images.
% to compare FORBES to the current best methods in the literature. 
These environments provide different challenges.
The Cartpole-Swingup task requires a long planning horizon and memorizing the state of the cart when it is out of view; Finger-Spinning includes contact dynamics between the finger and the object; Cheetah-Run exhibits high-dimensional state and action spaces; the Walker-Walk and Walker-Run are challenging because the robot has to learn to first stand up and then walk; Hopper Stand is based on a single-legged robot, which is sensitive to the reaction force on the ground and thus needs more accurate control.
% In all tasks, the only observations are third-person camera images of size $64 \times 64 \times 3$ pixels, the actions range from 1 to 12 dimensions, rewards range from 0 to 1, episodes last for 1000 steps, and have randomized initial states. 
As for baselines, we include the scores for A3C \citet{mnih2016asynchronous} with state inputs (1e9 steps),
D4PG \citet{barthmaron2018distributed} (1e9 steps), PlaNet \citep{hafner2019learning} (1e6 steps) and Dreamer \citet{hafner2019dream} with pixel inputs. All the scores of baselines are aligned with the ones reported in \citet{hafner2019dream} (see details in Appendix \ref{baselines}). We use $N=4$ trajectories.
The details of the implementations and hyperparameters can be found in Appendix \ref{sec:hparams}.

Our experiments empirically show that FORBES achieves superior performance and sample efficiency on challenging visual-motor control tasks.
As illustrated in Figure \ref{fig:mujoco-exp}, FORBES achieves higher scores than Dreamer \citep{hafner2019dream} in most of the tasks and achieves better performance than PlaNet \citep{hafner2019learning} within much fewer environment steps. See Appendix \ref{append-dmc-1m} for more results. We provide some insights into the results. As shown in Section \ref{exp:seqMNIST}, baselines with Gaussian assumptions may suffer from the mixed-up belief and empty belief issues, while FORBES can better capture the general belief states. Furthermore, multiple imagined trajectories can better utilize the diversity in the rollout. Therefore, the inner coherency within the model components allows the agent a better performance. We further discuss the role of multiple imagined trajectories and other components in the next section.
\vspace{-5pt}
\subsection{Ablation Study} 
\vspace{-5pt}
\begin{figure}[t]
    \centering
    \includegraphics[width=\linewidth]{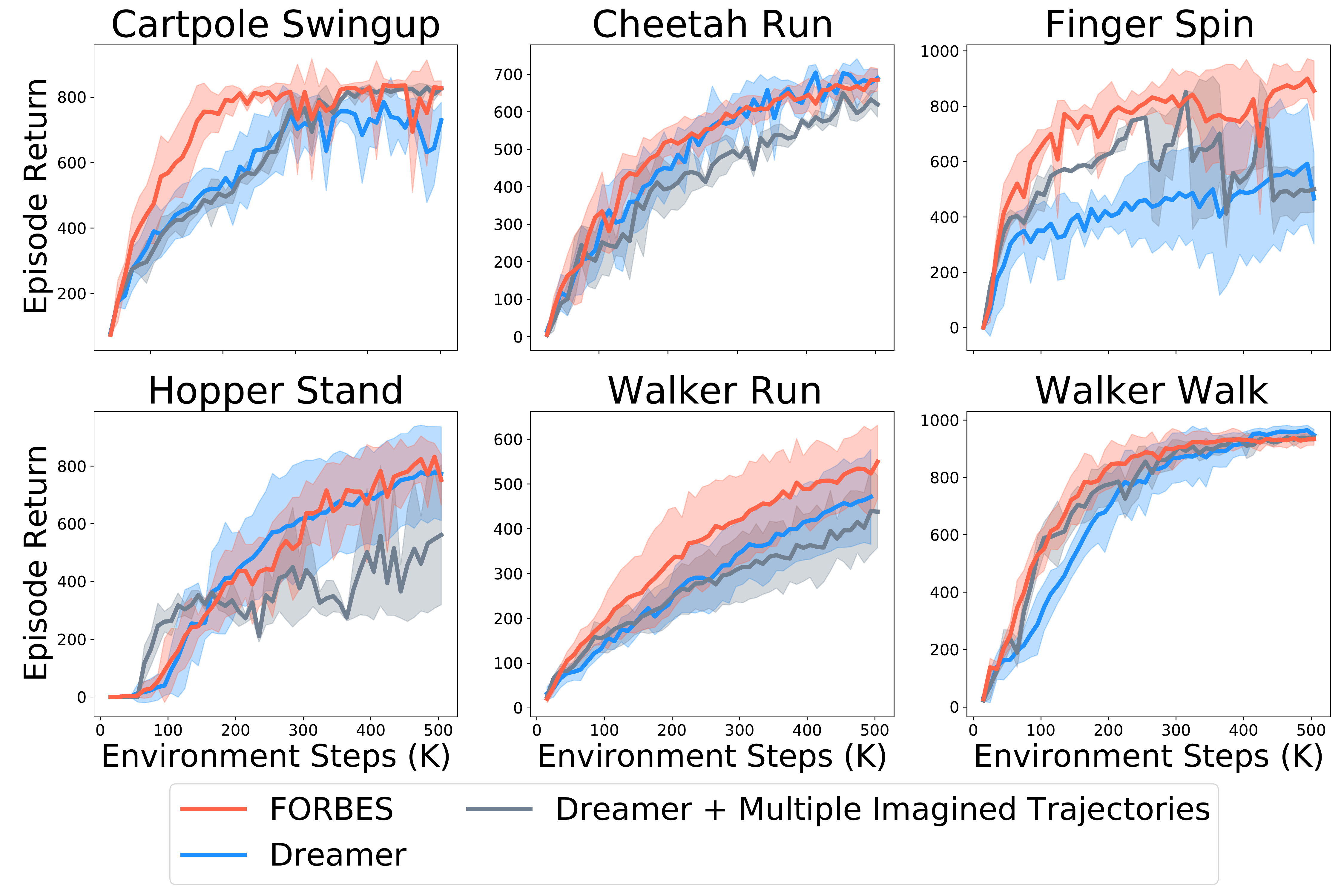}
    
    \centering
    \caption{Comparison of the performance between FORBES and Dreamer with multiple imagined trajectories.}
    \label{fig:ablation}
\vspace{-5pt}
\end{figure}

\label{exp:ablation}
In order to verify that the outperformance of FORBES is not simply due to increasing the number of imagined trajectories, we conducted an ablation study in this section. 
We compare FORBES with the ``Dreamer + multiple imagined trajectories" baseline by increasing the number of imagined trajectories in Dreamer to the same as in FORBES ($N=4$).
% Figure \ref{fig:ablation} shows the advantage of FORBES's utilization effect of multiple imagined trajectories compared to just increasing the number of imagined trajectories in Dreamer. Furthermore, 
As shown in Figure \ref{fig:ablation}, no consistent and obvious gain can be observed after increasing the number of trajectories to Dreamer. The agent gains slight improvements in two environments and suffers from slight performance loss on other tasks. 
This result indicates that increasing the number of imagined trajectories may only be effective when the agent can make diverse predictions as in FORBES.
The Gaussian assumptions lead to the lack of trajectory diversity, so that increasing the number of imagined trajectories will not effectively help.
Besides, Appendix \ref{append-ablation-n} compare different $N$ to illustrate the effect of multiple imagined trajectories. Appendix \ref{append-ablation-parameters} add parameters to baselines to illustrate the performance gain is not due to more parameters.

%% file: src/5-survey.tex
\vspace{-5pt}
\section{Related Work}
\vspace{-5pt}
\label{sec:relatedwork}
\textbf{POMDP:}  POMDP solving approaches can be divided into two categories based on whether their state, action and observation spaces are discrete or continuous. Discrete space POMDP solvers, in general, either approximate the value function using point-based methods \citep{kurniawati2008sarsop, shani2013survey} or using Monte-Carlo sampling in the belief space \citep{thrun1999monte, andrieu2002particle, silver2010monte, kurniawati2016online} to make the POMDP problem tractable. Monte Carlo algorithms like particle filters make it possible to handle POMDPs with continuous state space by maintaining sets of samples drawn from the belief states.
Other continuous space POMDP solvers often approximate the belief states as a distribution with few parameters (typically Gaussian) and solve the problem analytically either using gradients \citep{van2012motion, indelman2015planning} or using random sampling in the belief space \citep{agha2014firm, hollinger2014sampling}. 
However, most of the classical POMDP methods mentioned above are based on an accurately known dynamic model, which is a restricted assumption in many real world tasks.
%What's more, the aforementioned methods mainly focus on POMDPs with discrete state spaces. To track belief states with respect to continuous state space, another line of works use Monte Carlo algorithms like particle filter \citep{thrun1999monte, andrieu2002particle} to estimate belief states by maintaining sets of samples drawn from the belief states. 
More recently, 
\citet{nishiyama2012hilbert} proposes to solve the POMDP based on models defined in appropriate RKHSs, which represent probability distributions as embeddings in RKHSs. However, the embeddings are learned from training samples, and therefore this method requires access to samples from hidden states during training.

% Other approaches have also extended point-based methods to continuous domains \citep{shani2013survey}. Among continuous space POMDP solvers, \citet{agha2014firm} and \citet{hollinger2014sampling} have proposed sampling-based methods that can find effective solutions even in complex domains. However, most sampling-based methods suffer from the problem of obtaining sub-optimal solutions and can only be probabilistically optimal at best \citep{elbanhawi2014sampling}. 

\textbf{MBRL for visual-motor control:} 
Recent researches in model-based reinforcement learning (MBRL) for 
visual-motor control provides promising methods to solve POMDPs with high-dimensional continuous space and unknown models since visual-motor control tasks can be naturally modelled as POMDP problems.
%Classic model-based reinforcement learning approaches have focused on optimizing the policy in manually predefined low-dimensional state space  \citep{doya2002multiple}. 
% However, manually specifying a world model is difficult \citep{karkus2017qmdp,ma2020particle} in high dimentional visual control tasks. 
Learning effective latent dynamics models to solve challenging visual-motor control problems is becoming feasible through advances in deep generative modeling and latent variable models \citep{krishnan2015deep,karl2016deep,doerr2018probabilistic,buesing2018learning,ha2018world,han2019variational,hafner2019learning,hafner2019dream}. 
% World Model~\citep{ha2018world} builds generative neural network models of popular reinforcement learning environments by VAE.
Among which, the recurrent state-space model (RSSM) based methods \citep{hafner2019learning,hafner2019dream} provide a principled way to learn continuous latent belief states for POMDPs by variational inference and learns behaviours based on the belief states using model-based reinforcement learning, which achieves high performance on visual-motor control tasks.
% PlaNet \citep{hafner2019learning} proposes a recurrent state-space model (RSSM) by variational inference, which improves the effectiveness of historical information embedding and stochastic modelling. Dreamer~\citep{hafner2019dream} learns behaviours by propagating analytic gradients of learned state values back through long-horizon trajectories imagined by the RSSM. 
However, they assume the belief states obey diagonal Gaussian distributions. Such assumptions impose strong restrictions to belief inference and lead to limitations in practice, including mode collapse, posterior collapse and object vanishing in reconstruction\citep{bowman2015generating, salimans2015markov,okada2020dreaming}. In addition to the diagonal Gaussian distributions, \cite{tschiatschek2018variational} uses a Gaussian mixture to approximate the belief states. More recently, \cite{hafner2021mastering} proposes to approximate the belief states by assuming a discrete latent space and results in superior performance. However, our algorithm makes no assumption and has the capability to approach arbitrary continuous distribution according to the theoretical analysis. 
Other works like \cite{hausknecht2015deep, gregor2019shaping} use a vector-based representation of belief states. However, this deterministic representation prohibits the agent from consistently forecasting the future since the results of the reconstructed observation contain multimodality, and one can hardly keep the samples stay in the same mode across time.
Please check Appendix \ref{append-relatedworks} for more details.
A few works propose particle filter based methods that use samples to approximate the belief states~\citep{ma2020particle,igl2018deep}. However, particle filters are reported to experience the curse of dimensionality\cite{daum2003curse, cod2008} and therefore suffer from insufficient sample efficiency and performance \citet{lee2020stochastic}.
For a more detailed discussion of the related works, please refer to Appendix \ref{append-relatedworks}.

\textbf{Normalizing Flows:} Normalizing Flows (NF) are a family of generative models which produce tractable distributions with analytical density.
% where both sampling and density evaluation can be efficient and exact. 
For a transformation $f:\mathbf{R}^D \rightarrow \mathbf{R}^D$, the computational time cost of the log determinant is $\mathcal{O}(D^3)$. Thus most previous works choose to make the computation more tractable.
\cite{rezende2015variational, berg2019sylvester} propose to use restricted functional form of $f$.
% to exploit determinant identities. However, the inverse is hard to compute. 
Another choice is to force the Jacobian of $f$ to be lower triangular by using an autoregressive model \citep{kingma2016improved, papamakarios2018masked}.
% where the $i$-th dimension of the resulting variable depends on the first $i$ dimensions of the input.
These models usually excel at density estimation, but the inverse computation can be time-consuming.
\citet{dinh2014nice, dinh2017density, kingma2018glow} propose the coupling method to make the Jacobian triangular and ensure the forward and inverse can be computed with a single pass. 
% \citep{dinh2014nice} introduced affine normalizing flows with triangular Jacobians. \citep{dinh2016density} extend flows with masked convolutions, which allow for complex (non-autoregressive) dependence between the dimensions. In \citep{kingma2018glow}, $1 \times 1$ convolutions were proposed for improved image generation compared to \citep{dinh2016density}.  
% In \citep{huang2018neural} normalizing flows are auto-regressive and \citep{behrmann2018invertible} extend it to ResNet.
% \citep{lu2019structured} extended normalizing flows to model conditional distributions.
The applications of NF include image generation \citep{ho2019flow++,kingma2018glow}, video generation \citep{kumar2019videoflow} and  reinforcement learning \citep{mazoure2020leveraging,ward2019improving,touati2020randomized}.

%% file: src/6-conclusion.tex
\vspace{-5pt}
\section{Conclusion}
\vspace{-5pt}
% In this paper, we propose an algorithm for learning the belief state in POMDP. 
General continuous belief states inference is a crucial yet challenging problem in high-dimensional Partially Observable Markov Decision Process (POMDP) problems. In this paper, we propose the \textbf{F}l\textbf{O}w-based \textbf{R}ecurrent \textbf{BE}lief \textbf{S}tate model (FORBES)
that can learn general continuous belief states by incorporating normalizing flows into the variational inference framework and then
% incorporate FORBES into an RL framework to show that one can
effectively utilize the learned belief states in downstream RL tasks. 
We show that theoretically, our method can accurately learn the true belief states and 
we verify the effectiveness of our method in terms of both the quality of learned belief states and the final performance of our extended POMDP RL framework on two visual input environments.
The digit writing tasks demonstrate that our method can learn general belief states that enable precise and multi-modal predictions and high-quality reconstructions.
General belief inference plays a vital role in solving the POMDP, and our method paves a way towards it.  
In the future, we will explore further approaches to improve the accuracy of belief states inference and information seeking, such as combining contrastive learning and using advanced network architectures such as transformers to build normalizing flows.

%% file: src/7-sup.tex
\section{Details of affine coupling layer for normalizing flow}
\label{affine details}
In this section, we will introduce the details about the affine coupling layer \citep{dinh2017density}. 

In the forward function, we split the input $\mathbf{x}\in \mathbb{R}^{D}$ into two parts according to the dimension: $\mathbf{x}=[ \mathbf{x}_{1:k}, \mathbf{x}_{k+1:D} ]$. Then, we let the first part $\mathbf{x}_{1:k}$ stay identical, so that the first $k$ dimensions in the output $\mathbf{y} \in \mathbb{R}^{D}$ is $\mathbf{y}_{1:k}=\mathbf{x}_{1:k}$. After that, we use the identical part as the inputs to determine the transform parameters. In our case, we define two neural network $s,t: \mathbb{R}^k \rightarrow \mathbb{R}^{D-k}$, which stand for scale and translation functions. They receive $\mathbf{x}_{1:k}$ as inputs and output the affine parameters. As in \citep{dinh2017density}, the second part can be derived by:
\begin{equation}
    \mathbf{y}_{k+1:D} = \mathbf{x}_{k+1:D} \odot \exp(s(\mathbf{x}_{1:k})) + t(\mathbf{x}_{1_k})
\end{equation}
Finally, the output, $\mathbf{y}$ is the concatenation of the two parts: $\mathbf{y} = [\mathbf{y}_{1:k}, \mathbf{y}_{k+1:D}]$.

The affine coupling layer is an expressive transformation with easily-computed forward and reverse passes. The Jacobian of affine coupling layer is a triangular matrix, and its log determinant can also be efficiently computed.

\section{Hyper Parameters and implementation details}
\label{sec:hparams}

\paragraph{Network Architecture}

We use the convolutional and deconvolutional networks that are similar to Dreamer\citep{hafner2019dream}, a GRU \citep{cho2014learning} with $200$ units 
% as deterministic path 
in the dynamics model, and implement all other functions as two fully connected layers of size $200$ with ReLU activations. Base distributions in latent space are $30$-dimensional diagonal Gaussians with predicted mean and standard deviation.
As for the parameters network
%($\mathrm{NN}$ in Equation \ref{sup:arch})
, we use a residual network composed of one fully connected layer, one residual block, and one fully connected layer. 
The residual network receives $\mathbf{x}_{a}$ and $c$ as input. The input is first concatenated with the context and passed into the network. The residual block passes the input through two fully connected layers and returns the sum of the input and the output. Finally the last layer outputs the parameters 
%namely $\log \mathbf{s}$ and $\mathbf{t}$ in Equation \ref{sup:arch}.
and we use $5$ layers of affine coupling flows with a LU layer between them.

In our case, we use samples from the belief distribution as the inputs to the actor and value function as an approximation to the actor and value function with belief distribution as input. Calculating $V(b)$ needs to integrate through both the observation model and state transition model. Our approximation makes an assumption like in Qmdp, to avoid integrating through the observation model. 

    We use a GRU as the recurrent neural network to summary to temporal information. We assume an initial state $s_0$ to be a zero vector. After taking action $a_t$, we concatenate $a_t$ with the previous state $s_t$ and pass it through a small MLP to get $y_t = f(s_t, a_t)$, and use it as the input to the GRU: $h_{t+1}, z_{t+1} = GRU(h_t, y_t) $. We pass $z_{t+1}$ through an MLP to get the base prior belief distribution $p_0$ (mean and variance) and then we sample from $p_0$ and pass it through a sequence of Normalizing Flow to get a sample from $p_K$. For the posterior distribution, we first use a CNN as encoder to encode the observation $o_t$ into the feature $x_t$, and then concatenate $z_{t+1}$ and $x_t$ and pass them through an MLP to get the base posterior belief distribution $q_0$ and a sequence of Normalizing Flow. Similarly, we finally get a sample $s_{t+1}$ from $q_K$. 
    
    \paragraph{Training Details}
    
    We basically adopt the same data buffer updating strategy as in Dreamer \cite{hafner2019dream}. First, we use a small amount of $S$ seed episodes ($S=5$ in DMC experiments) with random actions to collect data. After that, we train the model for $C$ update steps ($C=100$ in DMC experiment) and conduct one additional episode to collect data with small Gaussian exploration noise added to the action. Algorithm 1 shows one update step in $C$ update steps. After $C$ update steps, we conduct one additional episode to collect data (this is not shown in Algorithm 1). When the agent interacts with the environment, we record the observations, actions, and rewards of the whole trajectory ($(o_t, a_t, r_t)_{t=1}^{T}$) and add it to data buffer $\mathcal{B}$.

\paragraph{Hyperparameters}

For DMControl tasks, we pre-process images by reducing the bit depth to 5 bits and draw batches of 50 sequences of length 50 to train the FORBES model, value model, and action model models using Adam \citep{kingma2014adam} with learning rates $\alpha_{0}=5\times10^{-4}$, $\alpha_{1}=8\times10^{-5}$, $\alpha_{2}=8\times10^{-5}$, respectively and scale down gradient norms that exceed $100$. We clip the KL regularizers in $\mathcal{J}_{Model}$ below $3.0$ free nats as in Dreamer and PlaNet. The imagination horizon is $H=15$ and the same trajectories are used to update both action and value models. We compute the TD-$\lambda$ targets with $\gamma=0.99$ and $\lambda=0.95$. As for multiple imagined trajectories, we choose $N=4$ across all environments.

For digit writing experiments in Section \ref{exp:seqMNIST}, we decrease the GRU hidden size to be $20$, let the base distributions be a $2$-dimensional diagonal Gaussian and only use $3$ layers of affine coupling flows. For the image processing, we simply divide the raw pixels by $255$ and subtract $0.5$ to make the inputs lie in $[-0.5, 0.5]$. 

\section{Extended information of Baselines}
\label{baselines}
For model-free baselines, we compare with D4PG \citep{barthmaron2018distributed}, a distributed extension of DDPG, and A3C \citep{mnih2016asynchronous}, the distributed actor-critic approach. D4PG is an improved variant of DDPG \citep{lillicrap2015ddpg} that uses distributed collection, distributional Q-learning, multi-step returns, and prioritized replay. We include the scores for D4PG with pixel inputs and A3C \citep{mnih2016asynchronous} with vector-wise state inputs from DMCcontrol. For model-based baselines, we use PlaNet \citep{hafner2019learning} and Dreamer \citep{hafner2019dream}, two state-of-the-art model-based RL. PlaNet \citep{hafner2019learning} selects actions via online planning without an action model and drastically improves over D4PG and A3C in data efficiency. Dreamer \citep{hafner2019dream} further improve the data efficiency by generating imaginary rollouts in the latent space.

% The model is trained using the Adam optimizer \citep{kingma2014adam} with a learning rate of $10^{-3}$, $\epsilon=10^{-4}$, and gradient clipping norm of $1000$ on batches of $B=50$ sequence chunks of length $L=50$. 

% We do not scale the KL divergence terms relatively to the reconstruction terms but grant the model $3$ free nats by clipping the divergence loss below this value. 
\newpage
\section{Further Discussion on Related Works}
\label{append-relatedworks}

This section further discusses the relationship between our work and some related works \citet{gregor2019shaping, hafner2019dream, hafner2021mastering}.

\begin{figure}[h]
    \centering
    \includegraphics[width=0.75\linewidth]{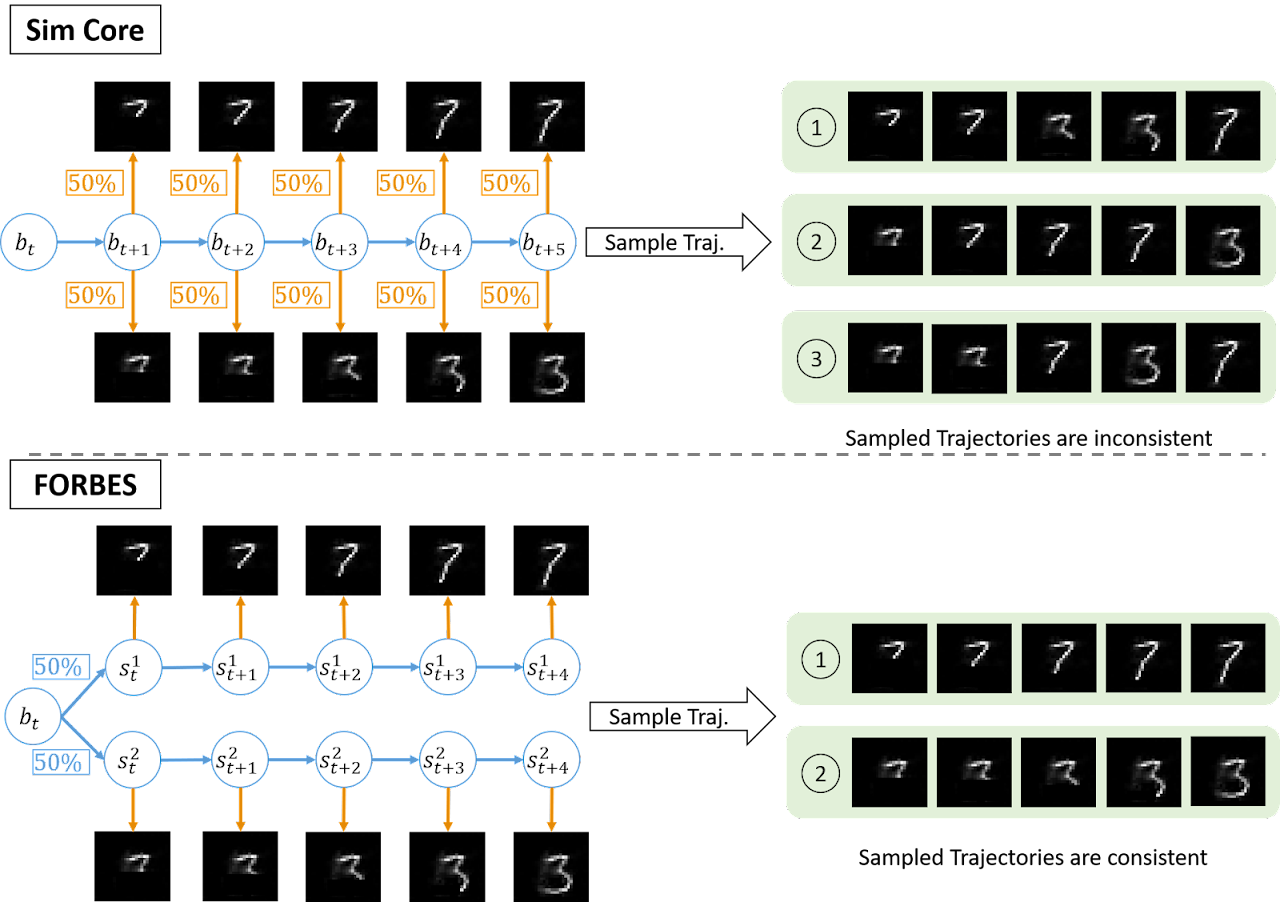}
    \centering
    \caption{Comparation of FORBES and SiMCore}
    \label{fig:sup-simcore}
\end{figure}

First of all, \citet{gregor2019shaping} proposes to use a flexible decoder to learn a compact vector representation of belief state and has promising results. Though it mentions using Normalizing Flow as a decoder, we believe it is orthogonal to our research. To summarize, we both aim to solve POMDP problems, hut our method has little in common with \cite{gregor2019shaping}, and our research directions are orthogonal: we use normalizing flows in different components for different purposes, which makes it unnecessary  for citing and comparing with \cite{gregor2019shaping}.

Specifically, our main contribution is to use Normalizing Flow to model accurate and flexible belief state distribution, and we prove its capability from theoretical and empirical perspectives.
\cite{gregor2019shaping} did not model belief state distribution. Instead, they model the belief as a single state vector, and the belief transition is also deterministic. Their Normalizing Flow is only used on image reconstruction. They use expressive generative models, including normalizing flow, to reconstruct the images conditioned on the simulated future state. "Using a convolutional DRAW outperforms flows for learning a model of complex environment dynamics" is reported in the context of reconstruction.

What's more, directly modeling the observation model from the belief in the form of a deterministic vector may have the following deficiencies:

First of all, it seems that SimCore cannot make consistent trajectory predictions by directly modeling the belief observation model. As shown in Figure \ref{fig:sup-simcore}. For instance, in the sequential MNIST setting, suppose the SimCore\cite{gregor2019shaping} is well trained, and we feed a beginning sequence which is the same as the last line in our Figure 4 into the SimCore. Assume it is equally possible to be ‘3’ or ‘7’, and the ConvDRAW will predict ‘3’ and ‘7’ each at 50\% probability at every time step. However, we cannot consistently sample the same category (‘3’ or ‘7’) in the same trajectory when we sample the future trajectory. However, since we explicitly model the state distribution for FORBES, we can first sample initial states and then rollout to sample multiple state trajectories, each covering a different category. This allows us to make diverse and consistent predictions.

Secondly, accurately obtaining the belief state is the main challenge in solving the POMDP. Dreamer \cite{hafner2021mastering} makes a strong isotropic Gaussian assumption to learn a continuous belief distribution, while Dreamer V2 \cite{hafner2021mastering} assumes discrete latent space. However, according to the theoretical analysis, our algorithm makes no assumption and can approach arbitrary continuous distribution. We believe that our methods can capture more subtle multimodal patterns without restricting the belief distribution to be discrete. This allows us to learn more general distribution (at least theoretically) and leaves great potential for future works.

To the best of our knowledge, we are the first to propose a normalizing flow based recurrent belief learning method to obtain the general continuous belief states in POMDP accurately. We provide theoretical analysis to illustrate that our algorithm has the potential of learning near perfectly accurate belief states. Through the sequential MNIST experiment, we empirically show the benefits of learning flexible belief distribution. Our method provides better reconstruction quality and can make multimodal future predictions. This flexible and accurate belief learning is essential for obtaining optimal solutions for POMDPs. 
As for the multiple imagined trajectories, we agree that the unimodal latent space leads to the lack of trajectory diversity, so that increasing the number of imagined trajectories will not effectively help. Our flexible belief distribution enables more accurate and multimodal future predictions by combining multiple imagined trajectories.

Therefore, we believe our proposed method is not merely a trivial combination of different components but a new framework for flexible and accurate belief distribution learning and POMDP RL with clear motivations and theoretical/empirical results.

% \subsection{Proofs of Theorem}\label{sec:proofs}

% % We adopt the defination of distributional universal approximator as in \citet{teshima2020couplingbased}.

\newpage
\section{An Ablation Study on the Number of Imagined Trajectories} \label{append-ablation-n}
\begin{figure}[ht]
    \centering
    \includegraphics[width=0.95\linewidth]{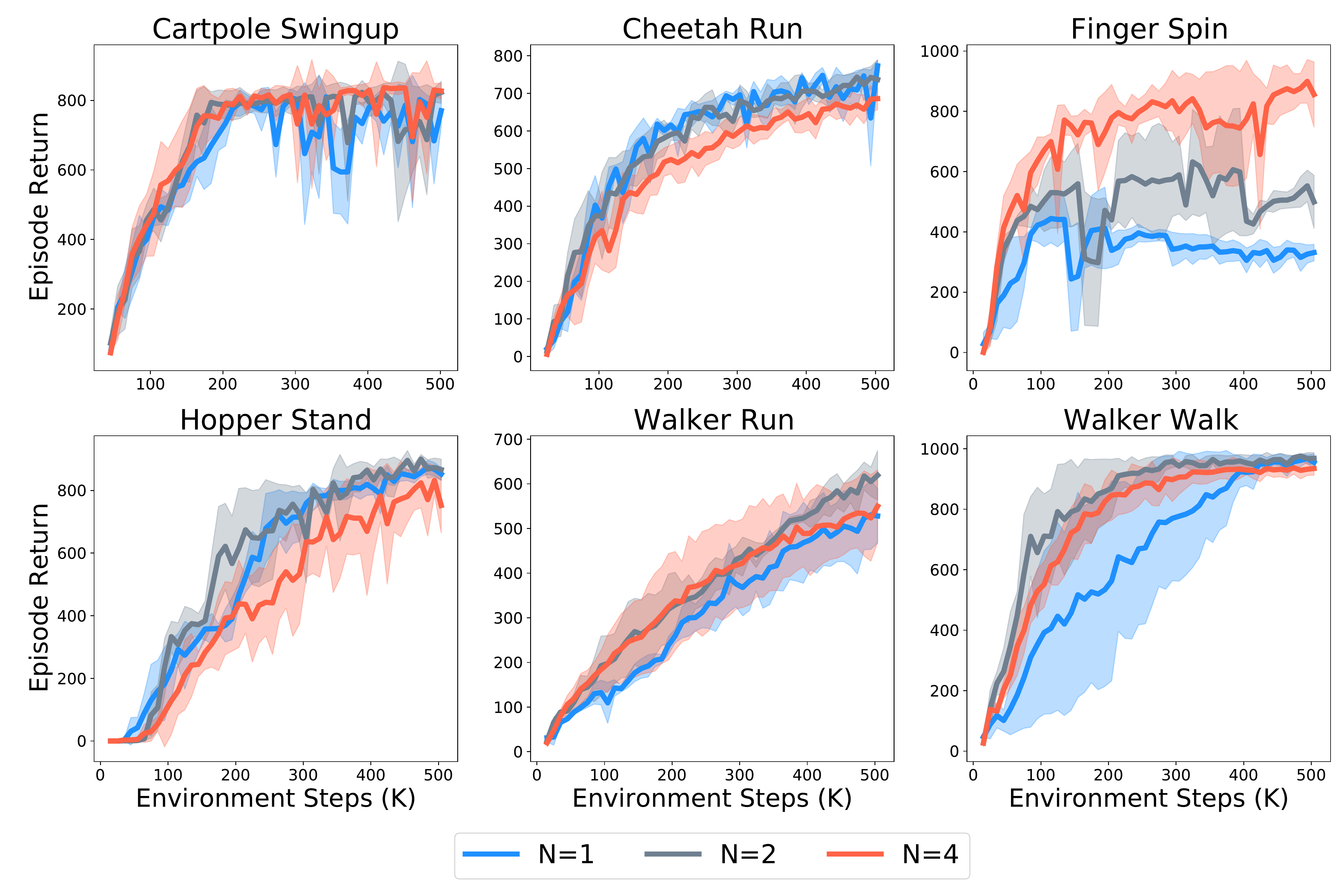}
    \centering
    \caption{An ablation study on the effect of different $N$ on DMC environments.}
    \label{fig:sup-abl-N}
\end{figure}
To show the effect of $N$, we adjust the number of imagined trajectories on some DMC environments. We choose $N=1, 2, 4$ and run 500K environment steps. We run $N=1,2$ with $3$ different seeds, and $N=4$ with $5$ different seeds (we use the main DMC experiment results, where $N=4$ here). The result shows that, in Finger Spin, the performance gain caused by multiple imagined trajectories is obvious. In finger spin, there are two objects and their interactions may result in complex locomotion patterns. When the environmental locomotion pattern itself is complex and flexible enough to incorporate diverse possibilities, then using FORBES allows the agent to make diverse predictions and using the multiple imagined trajectories technique will further exploit the advantages of FORBES. However, not all environments can show the advantages of multiple imaginations. In other environments, where there’s only one agent and its behavior is relatively unimodal, a larger $N$ does not effectively improve the performance, and different $N$ results in similar performances.

% \subsection{ELBO on DMC}

% \begin{figure}[ht]
%     \centering
%     \includegraphics[width=0.95\linewidth]{iclr2022/figs-rebuttal/DMC_1M.pdf}
    
%     \centering
%     \caption{The training curve on DMC environment for 1M environment steps.}
%     \label{fig:sup-DMC1M}
% \end{figure}

\newpage
\section{Extended Results on DMC} \label{append-dmc-1m}

\begin{figure}[ht]
    \centering
    \includegraphics[width=0.95\linewidth]{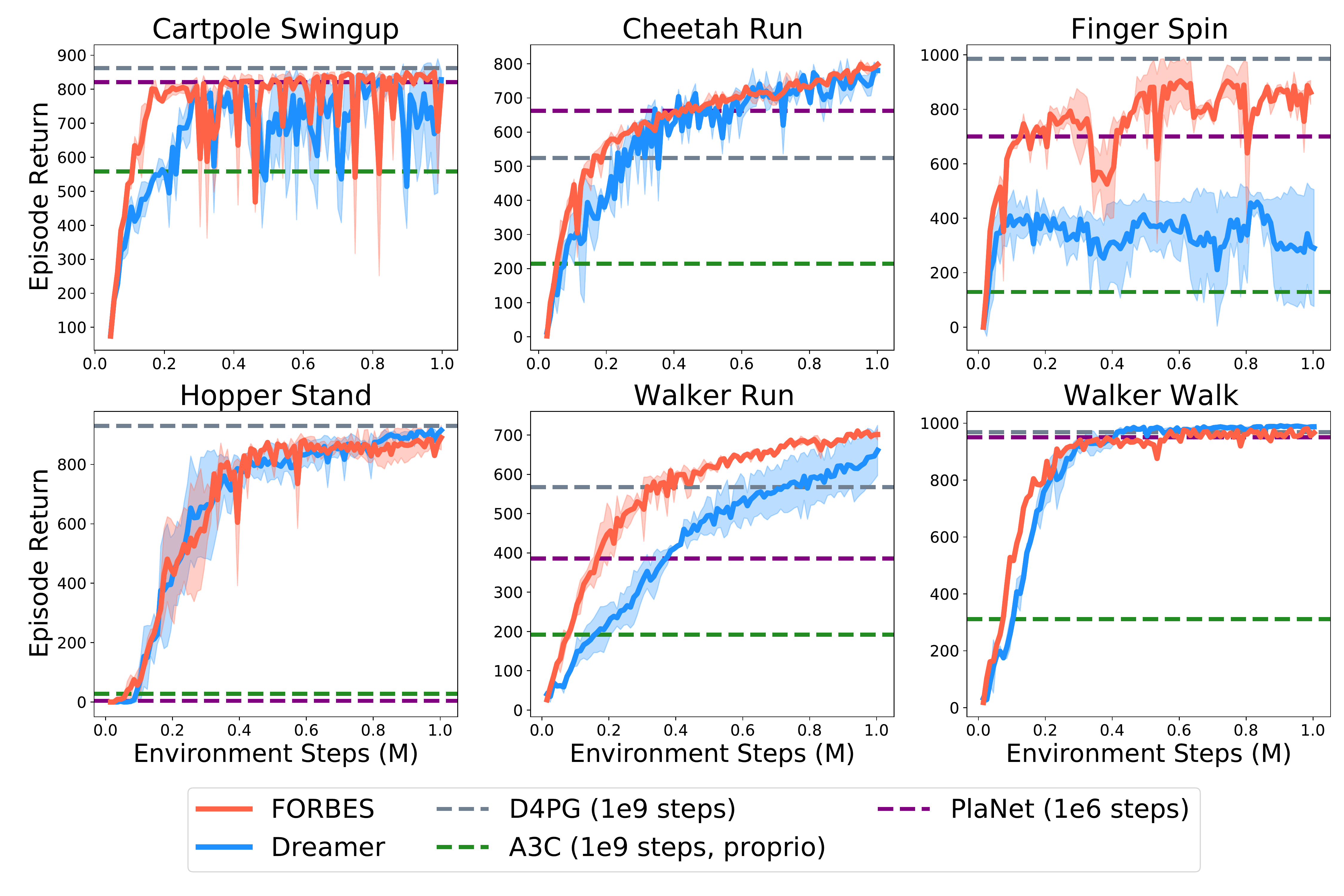}
    
    \centering
    \caption{The training curve on DMC environment for 1M environment steps.}
    \label{fig:sup-DMC1M}
\end{figure}

We run our algorithm for 1M environment steps and show the curve in Figure \ref{fig:sup-DMC1M}. We choose 1M environment steps because most of the curves have converged in most of the environments. FORBES achieves higher scores than Dreamer in most of the tasks.

\newpage
\section{An Ablation Study on the Model Parameters} \label{append-ablation-parameters}

\begin{figure}[ht]
    \centering
    \includegraphics[width=0.7\linewidth]{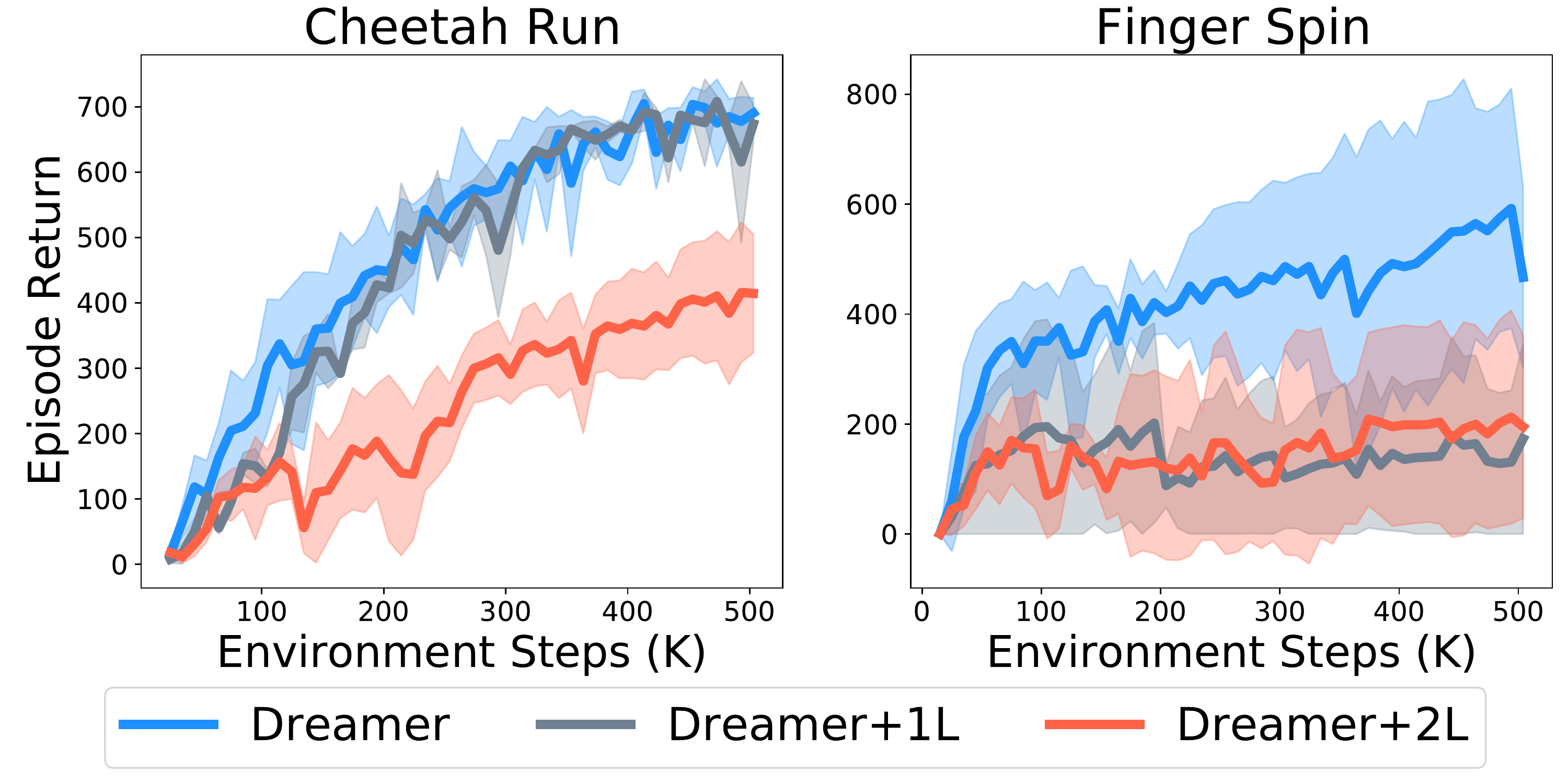}
    
    \centering
    \caption{An ablation study on the effect of adding parameters to Dreamer on two DMC environments.}
    \label{fig:sup-addparams}
\end{figure}

In this section, we show that having a flexible belief state distribution is the key to improving performance, rather than introducing more parameters. Having more parameters do not necessarily mean better performance. Increasing parameters may also make it difficult to converge and negatively affect the sample efficiency. 

We add an ablation study that adds more parameters to Dreamer to test the effectiveness of having more parameters. We add $1, 2$ hidden layer(s) to all the MLP in RSSM, and the result is shown in Figure \ref{fig:sup-addparams}.
The results show that simply adding parameters cannot improve the performance.

\newpage
\section{Comparison of ELBO on FORBES and RSSM on DMC }

\begin{figure}[ht]
    \centering
    \includegraphics[width=0.7\linewidth]{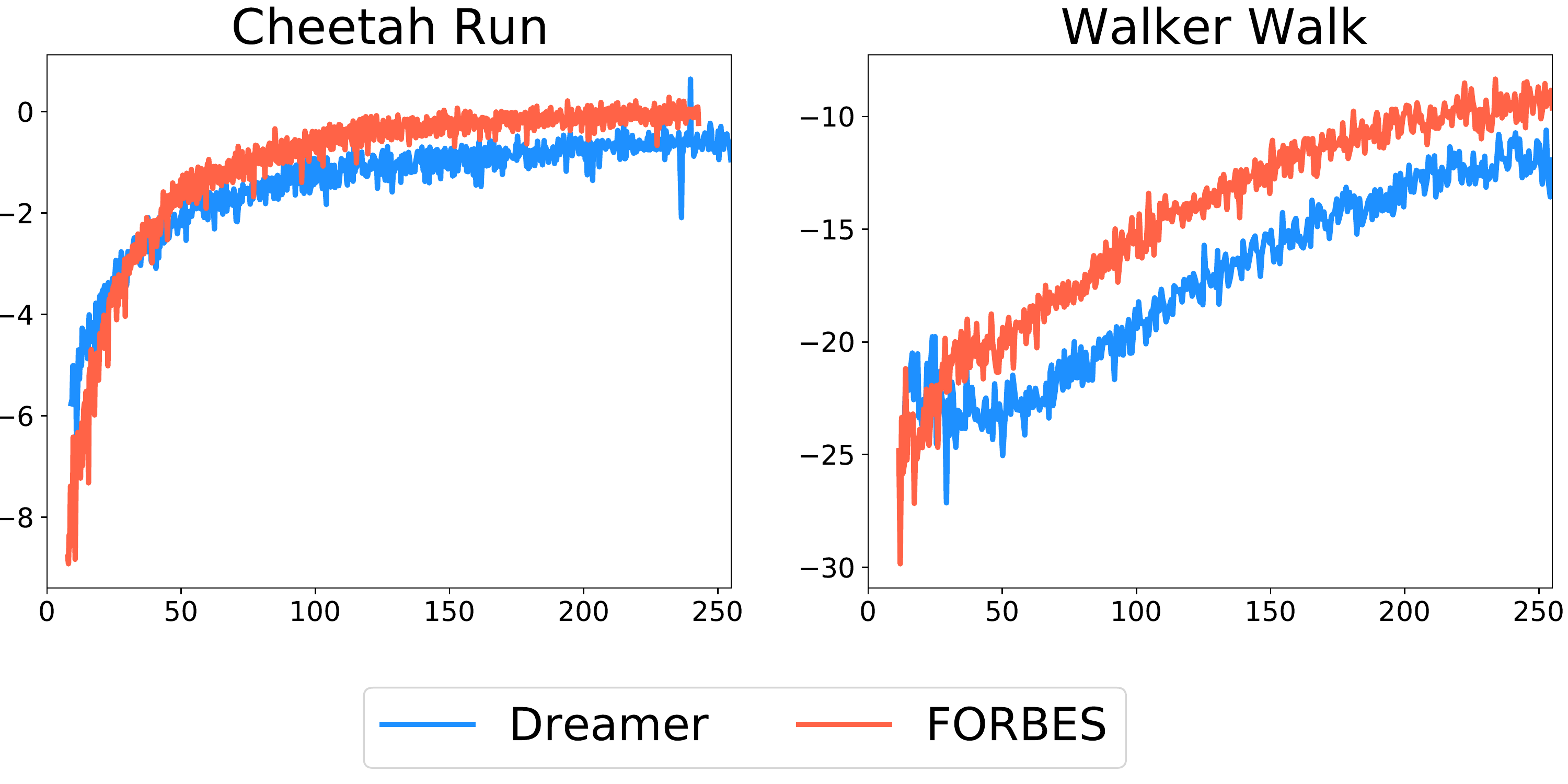}
    
    \centering
    \caption{The ELBO of FORBES and RSSM.}
    \label{fig:sup-DMCELBO}
\end{figure}

We provide the ELBO in DMC environments and FORBES in Figure \ref{fig:sup-DMCELBO}, and FORBES has higher ELBO.

\newpage
\section{Evidence Lower Bound Derivations}
\label{append-ELBO}

The variational bound for latent dynamics models
$p\left(o_{1: T}, s_{1: T} \mid a_{1: T}\right)=\prod_tp(s_t|s_{t-1},a_{t-1})p(o_t|s_t)$
 and a variational posterior $q\left(s_{1: T} \mid o_{1: T}, a_{1: T}\right)=\prod_{t} q\left(s_{t} \mid o_{\leq t}, a_{<t}\right)$ follows from importance weighting and Jensen's inequality as shown,

\begin{equation}
\begin{aligned}
\log p\left(o_{1: T} \mid a_{1: T}\right) &= \log \mathrm{E}_{p\left(s_{1: T} \mid a_{1: T}\right)}\left[\prod_{t=1}^{T} p\left(o_{t} \mid s_{t}\right)\right] \\
&=\log \mathrm{E}_{q\left(s_{1: T} \mid o_{1: T}, a_{1: T}\right)}\left[\prod_{t=1}^{T} p\left(o_{t} \mid s_{t}\right) p\left(s_{t} \mid s_{t-1}, a_{t-1}\right) / q\left(s_{t} \mid o_{\leq t}, a_{<t}\right)\right] \\
&\geq \mathrm{E}_{q\left(s_{1: T} \mid o_{1: T}, a_{1: T}\right)}\left[\sum_{t=1}^{T} \log p\left(o_{t} \mid s_{t}\right)+\log p\left(s_{t} \mid s_{t-1}, a_{t-1}\right)-\log q\left(s_{t} \mid o_{\leq t}, a_{<t}\right)\right].
\end{aligned}
\end{equation}

We use the same factorization of $q(s_{1:T}|\tau_t, o_t)$ in ELBO derivations and algorithm design as in \cite{hafner2019learning, hafner2019dream}.

% \begin{equation}
% \log p\left(o_{1: T}, r_{1: T} \mid a_{1: T}\right)-\mathcal{J}_{\text {Model }}=\sum_{t} D_{\mathrm{KL}}\left(q_{K}\left(s_{t} \mid \tau_{t}, o_{t}\right) \| p\left(s_{t} \mid s_{t-1}, a_{t-1}, o_{t}\right)\right)
% \end{equation}

% $D_{\mathrm{KL}}(q(s_{t} \mid \tau_{t}), o_{t} \| p(s_{t} \mid s_{t-1}, a_{t-1}, o_{t})) = \log p(o_{t}|a_{t}) + \int q(s_{t} \mid o_{t}) \log q(s_{t} \mid o_{t}) d s_{t}- \int q(s_{t} \mid o_{t}) \log [p(o_{t} \mid s_{t}) p(s_{t})] d s_{t}$

% $$

% $$

% $$
% =\int q(s_{t} \mid \tau_{t},o_{t}) \log \frac{q(s_{t} \mid \tau_{t},o_{t})}{p(s_{t} \mid s_{t-1},a_{t-1})} d s_{t}-\int q(s_{t} \mid \tau_{t},o_{t}) \log p(o_{t} \mid s_{t}) d s_{t}
% $$

% $$=D_{\mathrm{KL}}\left(q_{K}\left(s_{t} \mid \tau_{t}, o_{t}\right) \| p\left(s_{t} \mid s_{t-1}, a_{t-1}, o_{t}\right)\right) -\mathbb{E}_{q_{K}\left(s_{1: t} \mid \tau_{t}, o_{t}\right)} [\log p\left(o_{t} \mid s_{t}\right)]$$

\newpage
\section{Proofs of Theorem}\label{sec:proofs}
\textbf{Theorem 1:}The approximation error of the lower bound is
\begin{equation}
\nonumber
    \log p(o_{1:T}, r_{1:T}|a_{1:T}) - \mathcal{J}_{\mathrm{Model}} = \mathbb{E}_{q_K(s_{1:T}|\tau_T,o_T)} \left[ \sum_{t=1}^T  D_{\mathrm{KL}}(q(s_t | \tau_{t}, o_t) \| p(s_t\mid \tau_t, o_t)) \right]
\end{equation}
where $p(s_t\mid \tau_t, o_t)$ is the true posterior.

\textbf{Proof:}
\begin{equation}
\begin{split}
&D_{\mathrm{KL}}(q(s_{t} \mid \tau_{t},o_{t}) \| p\left(s_{t} \mid s_{t-1}, a_{t-1}, o_{t}\right)) \mid a_{1: T}\\
=&\int q(s_{t} \mid \tau_{t},o_{t}) \log \frac{q(s_{t} \mid \tau_{t},o_{t})}{p\left(s_{t} \mid s_{t-1}, a_{t-1}, o_{t}\right)} \mathrm{d} s_{t} \\
=&\int q(s_{t} \mid \tau_{t},o_{t}) \log \frac{q(s_{t} \mid \tau_{t},o_{t})}{\frac{p(s_{t} \mid s_{t-1},a_{t-1}) p\left(o_{t} \mid s_{t}\right)}{p(o_{t}\mid a_{1: T})}} \mathrm{d} s_{t} \\
=&\int q(s_{t} \mid \tau_{t},o_{t}) \log q(s_{t} \mid \tau_{t},o_{t}) \mathrm{d} s_{t}+\int q(s_{t} \mid \tau_{t},o_{t}) \log p(o_{t}\mid a_{1: T}) \mathrm{d} s_{t} \\
&\quad -\int q(s_{t} \mid \tau_{t},o_{t}) \log [p(s_{t} \mid s_{t-1},a_{t-1}) p(o_{t} \mid s_{t})] \mathrm{d} s_{t}\\
=& \log p(o_{t}\mid a_{1: T}) + \int q(s_{t} \mid \tau_{t},o_{t}) \log q(s_{t} \mid \tau_{t},o_{t}) \mathrm{d} s_{t} - \int q(s_{t} \mid \tau_{t},o_{t}) \log [p(s_{t} \mid s_{t-1},a_{t-1}) p(o_{t} \mid s_{t})] \mathrm{d} s_{t}\\
=& \log p(o_{t}\mid a_{1: T}) + \int q(s_{t} \mid \tau_{t},o_{t}) \log q(s_{t} \mid \tau_{t},o_{t}) \mathrm{d} s_{t} -\int q(s_{t} \mid \tau_{t},o_{t}) \log p(s_{t} \mid s_{t-1},a_{t-1}) \mathrm{d} s_{t}\\
& \quad -\int  q(s_{t} \mid \tau_{t},o_{t}) \log p(o_{t} \mid s_t) \mathrm{d} s_{t}\\
=& \log p(o_{t}\mid a_{1: T}) +\int q(s_{t} \mid \tau_{t},o_{t}) \log \frac{q(s_{t} \mid \tau_{t},o_{t})}{p(s_{t} \mid s_{t-1},a_{t-1})} \mathrm{d} s_{t}-\int q(s_{t} \mid \tau_{t},o_{t}) \log p(o_{t} \mid s_{t}) \mathrm{d} s_{t}\\
=& \log p(o_{t}\mid a_{1: T}) +D_{\mathrm{KL}}\left(q\left(s_{t} \mid \tau_{t}, o_{t}\right) \| p\left(s_{t} \mid s_{t-1}, a_{t-1}, o_{t}\right)\right) -\mathbb{E}_{q\left(s_{1: t} \mid \tau_{t}, o_{t}\right)} [\log p\left(o_{t} \mid s_{t}\right)]\\
\end{split}
\end{equation}

For a sequence from time 1 to T, we have
\begin{equation}
\begin{split}
&\sum_{t} D_{\mathrm{KL}}\left(q\left(s_{t} \mid \tau_{t}, o_{t}\right) \| p\left(s_{t} \mid s_{t-1}, a_{t-1}, o_{t}\right)\right)\\
= & \log p(o_{1: T}\mid a_{1: T})- \mathbb{E}_{q(s_{1:t}|\tau_t,o_t)} \left[ \sum_{t=1}^T (\log p(o_t | s_t)  - D_{\mathrm{KL}} (q(s_t|\tau_t,o_t) \| p(s_t |s_{t-1},a_{t-1}) )) \right] 
\label{eq:time-sequence}
\end{split}
\end{equation}

Then we can derive the Theorem 1 with \eqref{eq:time-sequence}:
\begin{equation}
\begin{split}
& \log p(o_{1: T},r_{1:T}\mid a_{1: T})\\
=& \mathbb{E}_{q_K(s_{1:T}|\tau_T,o_T)} \left[ \sum_{t} D_{\mathrm{KL}}\left(q\left(s_{t} \mid \tau_{t}, o_{t}\right) \| p\left(s_{t} \mid s_{t-1}, a_{t-1}, o_{t}\right)\right) \right]\\
&+\mathbb{E}_{q(s_{1:T}|\tau_T,o_T)} \left[ \sum_{t=1}^T (\log p(o_t | s_t) + \log p(r_t | s_t) - D_{\mathrm{KL}} (q(s_t|\tau_t,o_t) \| p(s_t |s_{t-1},a_{t-1}) )) \right]\\
=& \mathbb{E}_{q_K(s_{1:T}|\tau_T,o_T)} \left[ \sum_{t} D_{\mathrm{KL}}\left(q\left(s_{t} \mid \tau_{t}, o_{t}\right) \| p\left(s_{t} \mid s_{t-1}, a_{t-1}, o_{t}\right)\right) \right] + \mathcal{J}_{\mathrm{Model}}\\
=& \mathbb{E}_{q_K(s_{1:T}|\tau_T,o_T)} \left[\sum_{t} D_{\mathrm{KL}}\left(q\left(s_{t} \mid \tau_{t}, o_{t}\right) \| p\left(s_{t} \mid \tau_t, o_{t}\right)\right) \right] + \mathcal{J}_{\mathrm{Model}}
\end{split}
\end{equation}
where $p(s_t\mid s_{t-1},a_{t-1}, o_t)=p(s_t\mid \tau_t, o_t)$ given the sampled $s_{t-1}$ from $q(s_{1:t}|\tau_t,o_t)$.

\newpage
\section{More Results on Digit Writing Experiments}
\label{sup:digit-exp}
\begin{figure}[ht]
    \centering
    \includegraphics[width=1.0\linewidth]{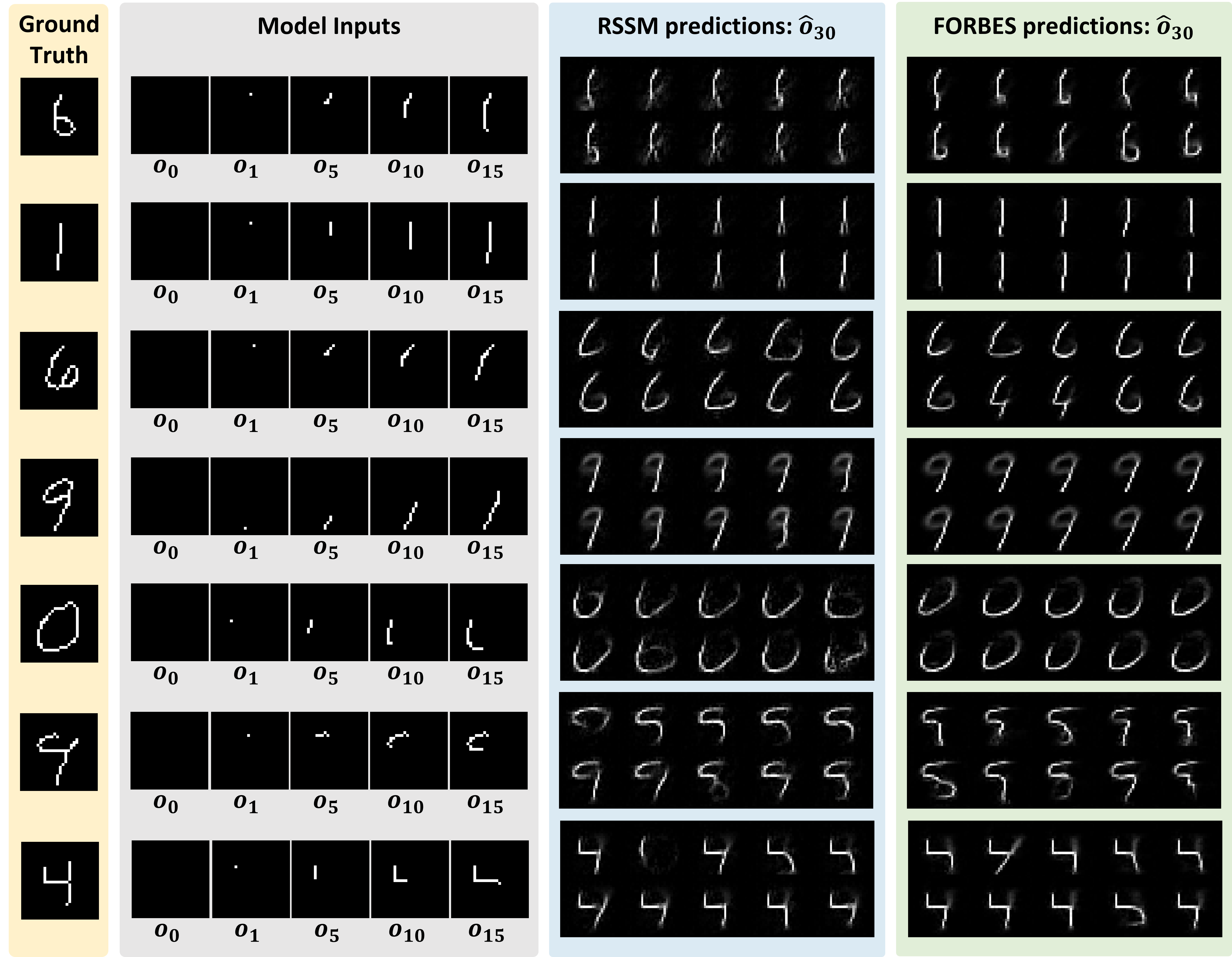}
    
    \centering
    \caption{Additional prediction results on sequential MNIST of two models.}
    \label{fig:sup-seq}
\end{figure}

In this section, we show more results of the predictions on the digit writing experiment in Figure \ref{fig:sup-seq}.

\newpage
\section{Reconstructions of the visual control tasks}

% \begin{figure}[ht]
%     \centering
%     \includegraphics[width=1.0\linewidth]{iclr2022/figs/sup-exp-mujoco-reconstruction/cheetah-run.png}
    
%     \centering
%     \caption{Reconstruction results of 10 frames on Cheetah Run.}
%     \label{fig:sup-mu-re}
% \end{figure}

\begin{figure}[hbpt]
    \centering
    \begin{subfigure}[h]{0.7\textwidth}
        \centering
        \includegraphics[width=1.0\textwidth]{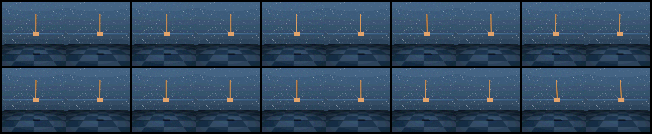}
        \caption{Cartpole Swing Up}
    \end{subfigure}%
    
    \begin{subfigure}[h]{0.7\textwidth}
        \centering  
        \includegraphics[width=1.0\textwidth]{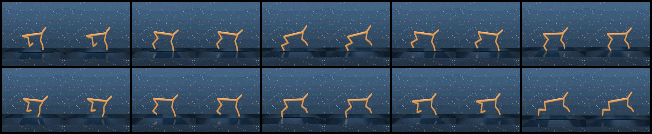}
        \caption{Cheetah Run}
    \end{subfigure}

    \begin{subfigure}[h]{0.7\textwidth}
        \centering
        \includegraphics[width=1.0\textwidth]{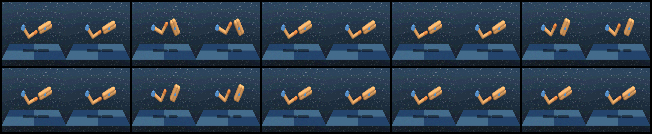}
        \caption{Finger Spin}
    \end{subfigure}
    
    \begin{subfigure}[h]{0.7\textwidth}
        \centering
        \includegraphics[width=1.0\textwidth]{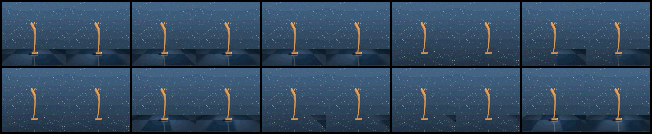}
        \caption{Hopper Stand}
    \end{subfigure}%
    
    \begin{subfigure}[h]{0.7\textwidth}
        \centering
        \includegraphics[width=1.0\textwidth]{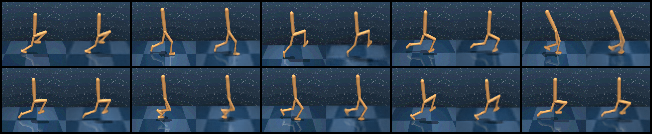}
        \caption{Walker Run}
    \end{subfigure}

    \begin{subfigure}[h]{0.7\textwidth}
        \centering
        \includegraphics[width=1.0\textwidth]{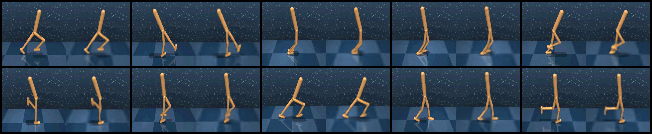}
        \caption{Walker Walk}
    \end{subfigure}
    \caption{The reconstruction results on of FORBES six environments from DeepMind Control Suite\citep{tassa2018deepmind}.}
    \label{fig:sup-mu-re}
\end{figure}

In this section, we show the reconstructions of the visual control tasks during the evaluating phase. 

For each environment, we use 10 frames. The left one is the original picture for each frame, and the right one is the reconstruction picture. The following results in Figure \ref{fig:sup-mu-re} show that FORBES can make high-quality reconstructions. The corresponding videos can be found in the supplementary material.